\documentclass[10pt,letterpaper,twocolumn]{article}
\usepackage{preprint}
\usepackage[hyphens]{url}  
\usepackage{graphicx} 
\usepackage{natbib}  
\usepackage{caption} 
\usepackage{multirow}
\usepackage{booktabs}
\usepackage{enumitem}
\usepackage{amsmath,amssymb}

\usepackage[hidelinks]{hyperref}
\hypersetup{
 pdftitle={Separating Capability from Confidence: Grounded Dual-State Calibration for GRPO-Trained Medical Vision-Language Models},
 pdfauthor={Yangyang Xie; Shangkun Li; Pengcheng Shi; Jiaqi Liu; Yun Gu; Xinglin Zhang},
 pdfsubject={Research preprint}
}
\setcitestyle{aysep={}}
\let\cite\citep
\title{Separating Capability from Confidence: Grounded Dual-State Calibration for GRPO-Trained Medical Vision-Language Models}
\author{
Yangyang Xie\textsuperscript{1,\(\ast\)},
Ke Hao\textsuperscript{1,\(\ast\)},
Jiaqi Liu\textsuperscript{2},
Yun Gu\textsuperscript{1,\(\dagger\)},
Xinglin Zhang\textsuperscript{2,\(\dagger\)}
}
\affiliations{
\textsuperscript{1}Shanghai Jiao Tong University, Shanghai, China\\
\textsuperscript{2}Medical Image Insights Co. Ltd., Shanghai, China\\[2pt]
\textsuperscript{\(\ast\)}Equal contribution.
\quad
Corresponding authors:
\href{mailto:yungu@ieee.org}{yungu@ieee.org}, \href{mailto:xinglinzh@gmail.com}{xinglinzh@gmail.com}}

\begin{document}

\maketitle

\begin{abstract}
Medical vision-language models (VLMs) require confidence that reflects both answer correctness and patient-specific visual evidence. Recent GRPO-based methods optimize verbalized confidence together with answer generation. However, this joint optimization may interfere with answer learning and drive confidence toward near-binary values. Verbalized confidence also provides no explicit assessment of visual support. We therefore separate capability learning from confidence estimation and propose \textbf{DualRead}. DualRead builds on the insight that reliability can be read from the actor's internal states at critical moments in the answering process. It freezes the GRPO-trained actor and combines pre-answer solvability with a post-answer assessment of the generated answer and its visual support. To further assess whether confidence reflects visual grounding, we introduce \textbf{Counterfactual Confidence Grounding AUC} (CCG-AUC). It measures whether confidence decreases when real-image substitution changes the actor from correct to incorrect. Across two VLM backbones and both in- and out-of-distribution medical VQA benchmarks, DualRead improves correctness discrimination and calibration over verbalized confidence while preserving answer accuracy. CCG-AUC reveals whether confidence responds to answer-relevant visual evidence rather than primarily to non-visual cues.
\end{abstract}

\section{Introduction}
\label{sec:introduction}

Large vision-language models (VLMs) show growing potential for medical visual question answering and clinical decision support. Yet even capable models make errors, and safe deployment requires identifying unreliable answers before they affect clinical decisions. Reliable confidence enables such a selective workflow: high-confidence cases can be handled automatically, while uncertain cases are deferred to clinicians~\citep{geifman2017selective}. This requires confidence to reflect the model's actual probability of correctness---a property that remains elusive in medical VQA, where overconfidence persists across model families and scales~\citep{byun2026overconfidence}.

A further requirement for medical VLMs is visual grounding. Clinical judgments rely on patient-specific evidence, whereas VLMs may produce correct or highly confident answers primarily from learned linguistic priors rather than the visual input. Reliable confidence should therefore indicate whether an answer is supported by evidence in the input image.
\begin{figure}[t]
\centering
\includegraphics[width=\linewidth]{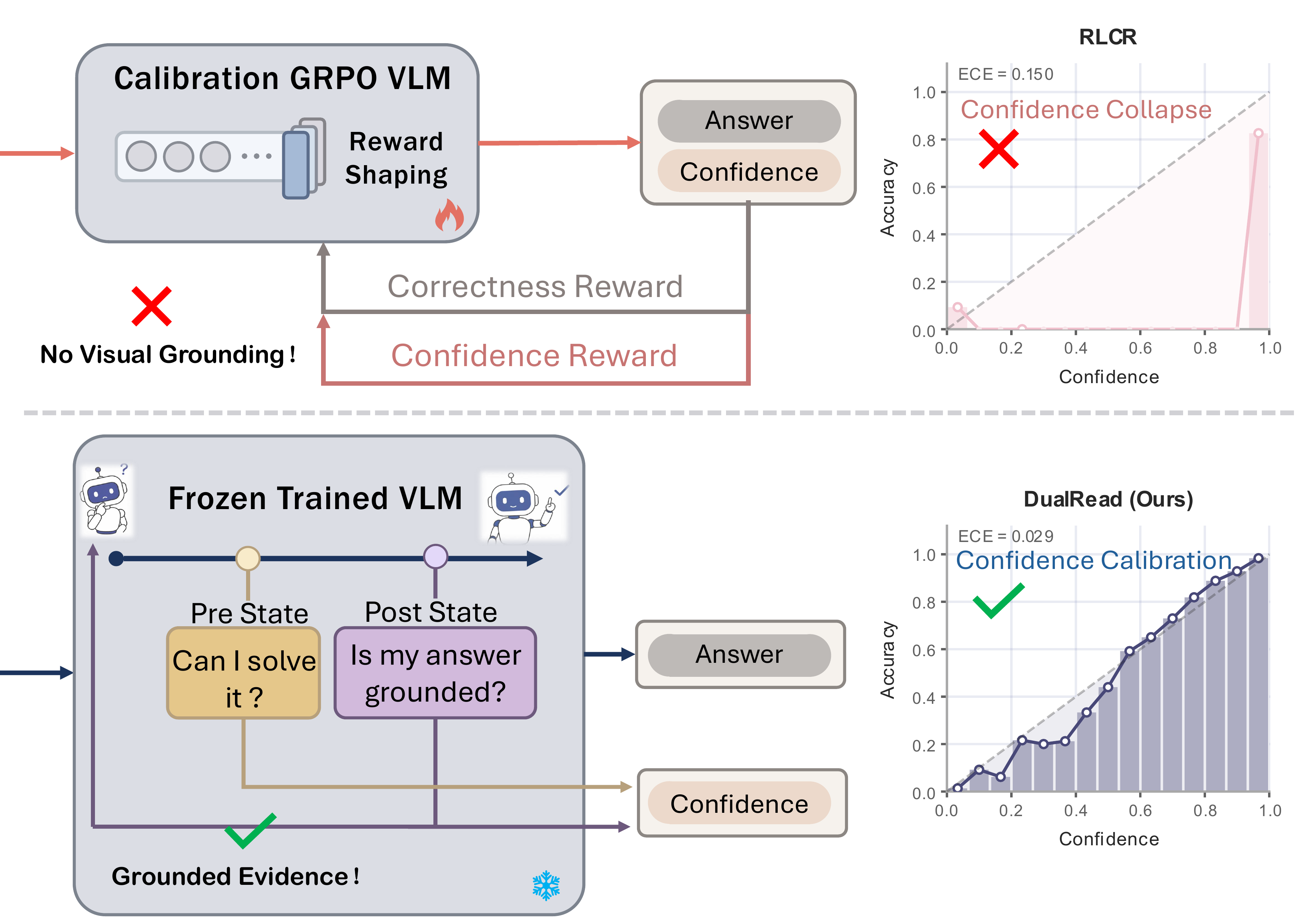}
\caption{\textbf{Motivation of DualRead.}
Reward-shaped GRPO calibration jointly optimizes answer correctness and
verbalized confidence, potentially causing confidence collapse (top). DualRead instead
keeps the RLVR actor frozen and estimates reliability along the same
natural generation trajectory (bottom), preserving the answer policy while
producing confidence that better tracks empirical accuracy.}
\label{fig:overview}
\end{figure}

Reinforcement learning from verifiable rewards (RLVR), particularly GRPO, has become an effective post-training paradigm for improving VLM answer capability. Recent efforts introduce verbalized confidence into the GRPO framework, aiming to improve answer correctness and confidence calibration jointly~\citep{ma2026dcpo}. However, the model output space is an imperfect interface for correctness confidence. A verbalized score such as ``90\% confident'' is another autoregressive action, affected by prompt wording, output format, and reward design. Optimizing this score jointly with the answer may interfere with correctness learning. Moreover, directly rewarding the score can polarize its distribution toward near-binary values (Figure~\ref{fig:overview}). These limitations motivate separating capability learning from confidence estimation.

We realize this separation with \textbf{DualRead}. The actor is trained only for answer capability with GRPO and is frozen after training. For confidence estimation, rather than asking the actor to report its own confidence, DualRead infers confidence by observing the internal states. Specifically, the actor first produces its natural response. The input and generated response are then replayed under teacher forcing in order to extract the required hidden states. DualRead focuses on two special times in this trajectory: immediately before response generation and after the answer is complete. The \emph{pre-answer solvability read} (PSR) uses the pre-answer state to estimate whether the actor can solve the input. The \emph{post-answer evidence read} (PER) reads the completed-answer state and incorporates visual evidence conditioned on the generated answer to assess whether the answer is supported by the image. The logits estimated by PSR and PER are calibrated separately and combined through dual-state fusion to obtain the final confidence probability. In this way, DualRead separates confidence estimation from capability learning while replacing subjective self-reporting with a state-based estimate.

To assess whether confidence is grounded in visual evidence, we introduce \textbf{Counterfactual Confidence Grounding AUC (CCG-AUC)}. For each question, the original image is replaced with a real hard negative that has a different reference answer. We focus on cases where the actor's answer changes from correct to incorrect after image replacement. CCG-AUC then evaluates whether confidence decreases accordingly by comparing the original image with its hard negative. This construction keeps the counterfactual inputs in distribution by using only real images. Fixing the question, actor, and decoding rule further isolates the effect of visual evidence on confidence. The metric is also scale-free and supports direct comparison across confidence interfaces on the same actor.



\section{Related Work}
\label{sec:related}

\paragraph{Reliability in medical VQA.}
Benchmarks such as SLAKE~\citep{liu2021slake}, VQA-RAD~\citep{lau2018vqarad}, PathVQA~\citep{he2020pathvqa}, and PMC-VQA~\citep{zhang2024pmcvqa} have driven rapid progress in medical VLMs, from instruction-tuned adaptations like LLaVA-Med~\citep{li2023llavamed} to medically pre-trained backbones like MedGemma~\citep{sellergren2026medgemma}. Reliability has received comparatively less attention: verbalized uncertainty in medical VQA remains poorly calibrated, and recent work improves it mainly through prompting and answer-set strategies~\citep{senoglu2026medvqa} or semantic perturbation~\citep{zhao2025csp}. These approaches all refine the model's \emph{self-reported} confidence.

\paragraph{RLVR and calibration under policy optimization.}
RLVR with GRPO~\citep{shao2024deepseekmath,guo2025deepseekr1} improves reasoning via outcome-verifiable rewards, but offers no supervision for deployment-time confidence---and RL post-training is known to harm calibration~\citep{damani2025rlcr,ma2026dcpo}. Recent calibration-aware variants incorporate confidence into the policy objective: RLCR adds a proper scoring rule~\citep{damani2025rlcr}, DCPO decouples confidence from reasoning gradients~\citep{ma2026dcpo}, and multimodal extensions include ranking-aware calibration~\citep{cui2026rac} and VL-Calibration~\citep{xiao2026vlcalibration}. Yet all express confidence through the generated token stream of a jointly optimized policy---the same output interface that suffers from prompt sensitivity and format artifacts. Beyond RL, confidence elicitation spans verbalized uncertainty~\citep{lin2022teaching,tian2023justask,xiong2024can}, $P(\mathrm{True})$ self-evaluation~\citep{kadavath2022know}, self-consistency~\citep{wang2023selfconsistency}, semantic uncertainty~\citep{kuhn2023semantic}, and hidden-state probes such as SAPLMA~\citep{azaria2023internal}---all of which, whether verbalized or probed, still operate through the output interface and are trained to predict correctness directly. What is missing, therefore, is a confidence estimator that is fully decoupled from the policy optimization that produces the answer, that reads internal states at semantically principled points along the generation trajectory rather than treating a single final state as a correctness classifier, and that explicitly accounts for whether the answer is visually grounded---not just whether it happens to be correct.

\paragraph{Evaluating visual grounding.}
That VQA models can answer from language priors while ignoring the image is well documented~\citep{goyal2017v,agrawal2018vqa,geirhos2020shortcut}, motivating counterfactual and contrastive constructions for training and evaluation~\citep{chen2021csst,gokhale2020mutant,wu2022similar,gardner2020contrast}. In the medical domain, \citet{zafar2026see} measure answer-side visual reliance with counterfactual images, and \citet{khanmohammadi2026bicr} rank confidence against blind (image-free) inputs. Existing protocols, however, target the \emph{answer} or rely on absent or corrupted images, which confounds grounding with novelty detection. CCG-AUC extends counterfactual evaluation to the \emph{confidence score} itself, using real in-distribution hard negatives and restricting to cases where the image decisively determines correctness. The two repairs are also complementary: \citet{zafar2026see} improve grounding by retraining the actor with a hard-negative contrastive objective, which changes its answers, whereas DualRead leaves the actor and its answers exactly unchanged and repairs only the confidence gap.

\section{DualRead Confidence Calibration}
\label{sec:method}

\begin{figure*}[t]
    \centering
    \includegraphics[width=\linewidth]{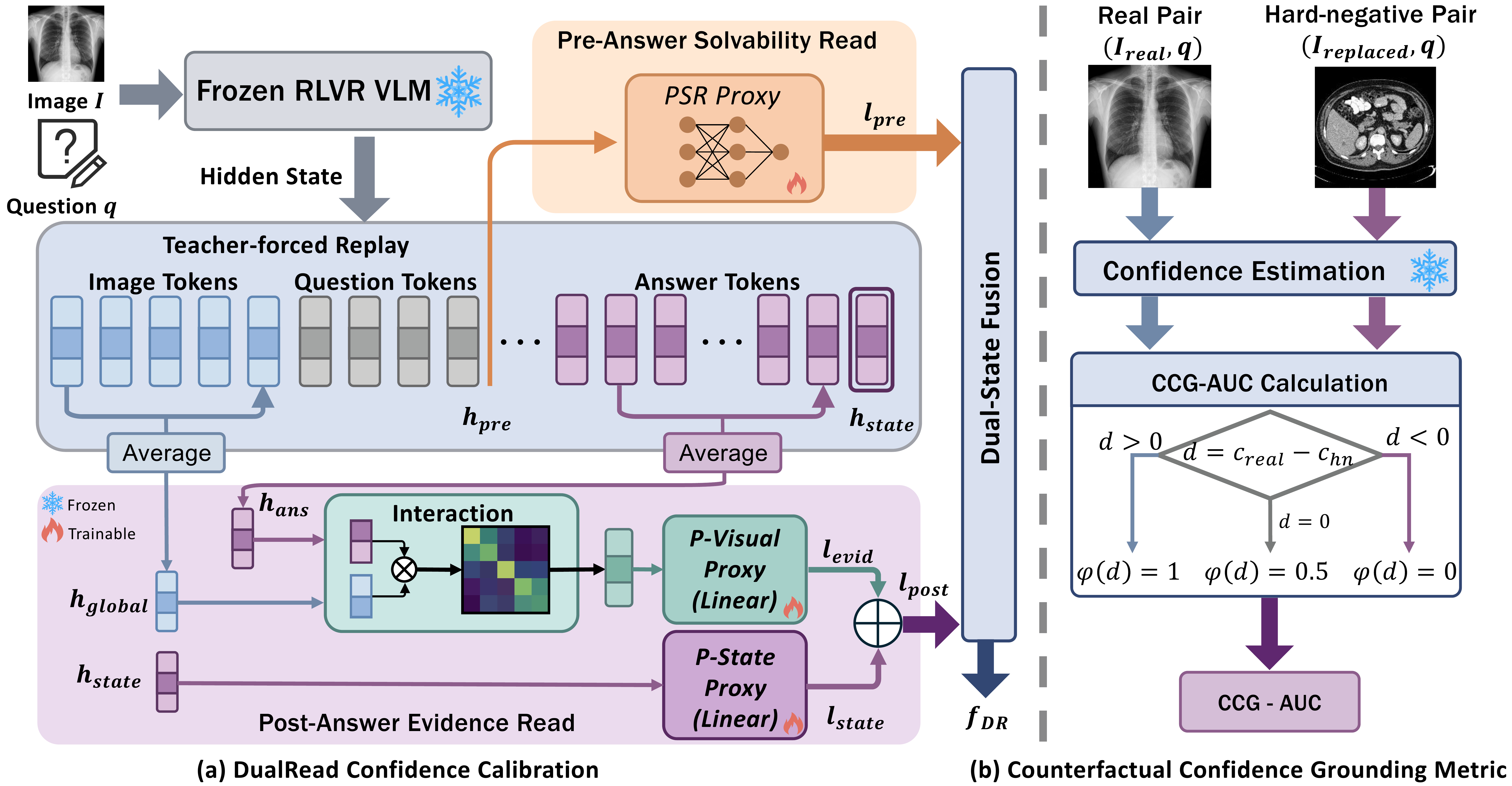}
    \caption{\textbf{Overview of our framework.} (a) DualRead estimates confidence from pre- and post-answer states of a frozen GRPO-trained VLM. (b) CCG-AUC evaluates confidence grounding using real and hard-negative image pairs sharing the same question.}   
    \label{fig:method}
    \end{figure*}


Let $I_i$, $q_i$, and $y_i^{*}$ denote a medical image, its associated question, and the reference answer. We train an actor $\pi_\theta$ with RLVR using GRPO~\citep{shao2024deepseekmath}. For each input $x_i=(I_i,q_i)$, the actor samples a group of $G$ responses $\{\hat y_{i,g}\}_{g=1}^{G}$. Each response receives a verifiable reward based on answer correctness $z_{i,g}\in\{0,1\}$ and output format. GRPO derives group-relative advantages from these rewards and optimizes a clipped policy objective, increasing the likelihood of responses that perform better within the group. This produces the trained actor $\pi_{\theta^{*}}$.

The resulting actor improves answer generation, while the correctness probability of each generated response remains unspecified. For a response $\hat y_i$, DualRead produces a confidence probability $p_i^{\mathrm{DR}}\in[0,1]$ that estimates its likelihood of being correct. A calibrated estimate should satisfy $\Pr(z_i=1\mid p_i^{\mathrm{DR}}=p)\approx p$, where $z_i$ indicates whether $\hat y_i$ matches the reference answer. Our central design choice is to learn this confidence outside policy optimization. Accordingly, we freeze the GRPO-trained actor $\pi_{\theta^{*}}$ and fit a separate confidence proxy to its internal states. Reliability supervision updates only the confidence proxy, with $\nabla_{\theta^{*}}\mathcal{L}_{\mathrm{rel}}=0$, leaving the actor and its answer policy unchanged.

\subsection{Natural-response Replay}
For each training example, the frozen actor first produces its natural response $\hat y_i$.  We parse this response and assign the realized-correctness label $z_i=\mathbf{1}[\operatorname{Correct}(\hat y_i,y_i^{*})]$, which provides supervision for the confidence proxies.  The original input and the natural response are then fed back into the frozen actor for a teacher-forced replay, allowing us to revisit the internal states.  During replay, DualRead extracts final-layer hidden states at two stages: the state immediately before response generation and the states that provide a retrospective view after the answer is complete.  

\subsection{Dual-state Confidence Proxies}
\label{sec:dualread}

DualRead instantiates the two replay views as complementary confidence proxies (Figure~\ref{fig:method}).  The \emph{pre-answer solvability read} estimates whether the input is solvable by the current actor before any response token is produced.  The \emph{post-answer evidence read} evaluates the particular answer that the actor actually committed to. 


\subsubsection{Pre-Answer Solvability Read (PSR)}
\label{sec:pre_read}

Let $H_{\theta^{*}}^{(L)}(I_i,q_i,\hat y_i)$ denote the sequence of $D$-dimensional hidden states produced by the final layer $L$ of the frozen actor during replay.  PSR extracts the state at the last non-padding prompt position, immediately before the first response token: $h_i^{\mathrm{pre}}=H_{\theta^{*}}^{(L)}(I_i,q_i,\hat y_i)_{t_i^{\mathrm{prompt}}}\in\mathbb{R}^{D}$, where $t_i^{\mathrm{prompt}}$ indexes this position.  By causal masking, $h_i^{\mathrm{pre}}$ can access only the input image $I_i$ and question $q_i$.  The state is therefore independent of the wording, length, and fluency of the realized response, and captures the actor's internal representation immediately before answering.  A lightweight MLP, termed the \emph{PSR proxy}, maps this state to a correctness logit, $\ell_i^{\mathrm{pre}}=f_{\mathrm{PSR}}(h_i^{\mathrm{pre}})$.  We train it using $z_i$ with class-weighted binary cross-entropy.  The proxy thus models the actor-specific relationship between its pre-answer state and the correctness of its subsequent response.  PSR is optimized independently and remains fixed when fitting PER.

\subsubsection{Post-Answer Evidence Read (PER)}
\label{sec:post_read}

Complementary to PSR, PER operates after the answer is complete.  It contains two components: one reads the actor's post-answer state, while the other evaluates the answer against the visual evidence.
\begin{itemize}[
    leftmargin=1.5em,
    labelsep=0.4em
]
\item {Answer commitment state (P-State).}
For the post-answer state, we extract the final-layer hidden state at the closing position $t_i^{\mathrm{close}}$ of answer tokens: $h_i^{\mathrm{close}}=H_{\theta^{*}}^{(L)}(I_i,q_i,\hat y_i)_{t_i^{\mathrm{close}}}\in\mathbb{R}^{D}$.  This position summarizes the image, question, generated reasoning, and complete response, providing a retrospective read after the answer has been formed.  P-State applies a linear confidence proxy to obtain the post-answer logit, $\ell_i^{\mathrm{state}}=f_{\mathrm{state}}(h_i^{\mathrm{close}})$.
    \item {Answer-conditioned visual evidence (P-Visual).}
P-Visual first extracts answer and image states, then applies answer-conditioned visual pooling to summarize the image evidence relevant to the generated answer.  We denote the token positions that form the answer by $A_i$ and summarize its content by averaging their final-layer states,
\begin{equation}
h_i^{\mathrm{ans}}
=
\frac{1}{|A_i|}
\sum_{t\in A_i}
H_{\theta^{*}}^{(L)}(I_i,q_i,\hat y_i)_t.
\label{eq:answer_state}
\end{equation}
Let $v_{i,1},\ldots,v_{i,M_i}\in\mathbb{R}^{D}$ be the final-layer states at the image-token positions.  Their global mean is
\begin{equation}
h_i^{\mathrm{global}}
=
\frac{1}{M_i}\sum_{j=1}^{M_i}v_{i,j}.
\end{equation}
We obtain the answer-conditioned evidence state through cosine-similarity pooling:
\begin{align}
\alpha_{i,j}
&=
\frac{
\exp\!\left(
\cos(h_i^{\mathrm{ans}},v_{i,j})/\kappa
\right)
}{
\sum_{k=1}^{M_i}
\exp\!\left(
\cos(h_i^{\mathrm{ans}},v_{i,k})/\kappa
\right)
},
\\
h_i^{\mathrm{evid}}
&=
\sum_{j=1}^{M_i}\alpha_{i,j}v_{i,j},
\label{eq:evidence_pool}
\end{align}
where $\kappa$ is a fixed pooling temperature.  This parameter-free operation summarizes image-token states that are representationally aligned with the answer.
We form the visual interaction block:
\begin{equation}
\phi_i^{\mathrm{evid}}
=
\left[
h_i^{\mathrm{global}};
h_i^{\mathrm{evid}};
h_i^{\mathrm{ans}}\odot h_i^{\mathrm{evid}};
\left|h_i^{\mathrm{ans}}-h_i^{\mathrm{evid}}\right|
\right]
.
\label{eq:visual_block}
\end{equation}
The interaction block is passed through the P-Visual confidence proxy $f_{\mathrm{visual}}$ and connected residually to the answer-state logit.  The final post-answer logit is
\begin{equation}
\ell_i^{\mathrm{post}}
=
\ell_i^{\mathrm{state}}
+
f_{\mathrm{visual}}(\phi_i^{\mathrm{evid}}).
\label{eq:post_logit}
\end{equation}
\end{itemize}

The proxies $f_{\mathrm{state}}$ and $f_{\mathrm{visual}}$ are trained sequentially using $z_i$ with binary cross-entropy and $\ell_2$ regularization.  P-State is fitted first and then frozen while P-Visual learns the residual correction. We exclude decoder likelihood and entropy, as these signals are more closely tied to the fluency and local predictability of the output trajectory than to its visual support.

\subsection{Calibration and Dual-State Fusion}
\label{sec:fusion}

PSR and PER produce logits at different stages of the response and may operate at different confidence scales.  We calibrate them separately using temperature scaling, yielding
\begin{equation}
p_i^{\mathrm{pre}}=\sigma\!\left(\ell_i^{\mathrm{pre}}/T_{\mathrm{pre}}\right),
\qquad
p_i^{\mathrm{post}}=\sigma\!\left(\ell_i^{\mathrm{post}}/T_{\mathrm{post}}\right).
\label{eq:branch_calibration}
\end{equation}
where each scalar temperature is selected by minimizing binary cross-entropy while the actor and confidence proxies remain fixed.  DualRead then combines the calibrated pre- and post-answer confidence using an equal-weight average in log-odds space:
\begin{equation}
p_i^{\mathrm{DR}}
=
\sigma\!\left[
\frac{1}{2}\operatorname{logit}(p_i^{\mathrm{pre}})
+
\frac{1}{2}\operatorname{logit}(p_i^{\mathrm{post}})
\right].
\label{eq:dualread_fusion}
\end{equation}
The resulting $p_i^{\mathrm{DR}}$ is the final confidence estimate for $\hat y_i$.  No parameters are learned during fusion, and the actor and its generated answer remain unchanged.

\section{Counterfactual Confidence Grounding Metric}
\label{sec:ccg}

Calibration alone does not show whether confidence responds to the image.  A confidence estimate may reflect question difficulty or language priors without tracking the visual evidence relevant to the answer.  We therefore introduce \textbf{Counterfactual Confidence Grounding AUC (CCG-AUC)}, which measures whether confidence decreases when answer-relevant visual evidence is replaced.

The counterfactual should remain within the data distribution.  Blanking or removing the image creates an artificial input and provides no valid counterfactual answer.  Instead, we replace the original image with a real image paired with the same question but a different reference answer.  This produces an in-distribution counterfactual in which the image is relevant to the expected answer.

For each item $(I_i,q_i,y_i^*)$, we select a real hard-negative image $I_i^{\mathrm{hn}}$ associated with the same question and a different reference answer.  Items with a valid pairing form the eligible set $\mathcal P$.  Using the same frozen actor and decoding rule, we generate an answer for both $(I_i,q_i)$ and $(I_i^{\mathrm{hn}},q_i)$.  This yields correctness labels $z_i^{\mathrm{real}}$ and $z_i^{\mathrm{hn}}$, evaluated with their respective reference answers, and confidence estimates $c_i^{\mathrm{real}}$ and $c_i^{\mathrm{hn}}$.  We focus on the decisive set
\begin{equation}
\mathcal E
=
\left\{i\in\mathcal P:
z_i^{\mathrm{real}}=1
\ \wedge\
z_i^{\mathrm{hn}}=0
\right\},
\label{eq:ccg_set}
\end{equation}
where image replacement changes the actor from correct to incorrect.  Define $\varphi(d)=\mathbf 1[d>0]+\tfrac12\mathbf 1[d=0]$.  CCG-AUC is then
\begin{equation}
\operatorname{CCG\mbox{-}AUC}
=
\frac{1}{|\mathcal E|}
\sum_{i\in\mathcal E}
\varphi\!\left(c_i^{\mathrm{real}}-c_i^{\mathrm{hn}}\right).
\label{eq:ccg_auc}
\end{equation}
The metric assigns $1$ when confidence is higher for the original image, $0$ when it is higher for the hard negative, and $0.5$ for a tie.  Thus, $0.5$ indicates no directional response, while $1.0$ indicates that confidence always decreases after decisive image replacement. It does not measure pixel-level localization or rationale faithfulness.  

We refer to Eq.~\eqref{eq:ccg_auc} as \textbf{Natural CCG-AUC}, since the
actor generates its response naturally for each image and may therefore
change its answer after image replacement. To control for answer changes,
let
$a_i^{\mathrm{real}}=\operatorname{canon}(\hat y_i^{\mathrm{real}})$ and
$a_i^{\mathrm{hn}}=\operatorname{canon}(\hat y_i^{\mathrm{hn}})$ denote the
canonical actor answers. We define the strict decisive set as
\begin{equation}
\mathcal E_{\mathrm{strict}}
=
\left\{
i\in\mathcal E:
a_i^{\mathrm{real}}=a_i^{\mathrm{hn}}
\neq\varnothing
\right\}.
\end{equation}
\textbf{Strict CCG-AUC} applies Eq.~\eqref{eq:ccg_auc} over
$\mathcal E_{\mathrm{strict}}$. It therefore evaluates whether confidence
responds to image evidence while holding the canonical answer fixed.


\section{Experiments and Results}
\begin{table*}[!t]
\centering
\captionsetup{width=\textwidth}
\caption{Reliability and confidence-grounding results on SLAKE and six OOD
medical VQA datasets. OOD Acc./AUROC are equal-domain averages over all six
datasets. Confidence proxies from GRPO (RLVR) to DualRead share
the same frozen actor and thus identical Acc. Gray and blue denote Base and
DualRead, respectively; bold indicates the best non-Acc. result.}
\label{tab:main}
\definecolor{MergedBaseGray}{RGB}{242,242,242}
\definecolor{MergedOursBlue}{RGB}{226,240,250}
\newcommand{\mergedms}[2]{\ensuremath{#1_{\pm\!#2}}}
\newcommand{\mergedbest}[2]{\ensuremath{\mathbf{#1}_{\pm\!#2}}}
\newlength{\mergeduparrowlen}
\newlength{\mergeddownarrowlen}
\setlength{\mergeduparrowlen}{20.6pt}
\setlength{\mergeddownarrowlen}{40.4pt}

\newcommand{\mergedsharedacc}[1]{%
  \begingroup
  \setbox0=\hbox{#1}%
  \makebox[0pt][l]{%
    \kern.5\wd0\kern-.8pt
    \makebox[0pt][c]{%
      \raisebox{\ht0}[0pt][0pt]{%
        \rotatebox{90}{%
          $\xrightarrow{\hspace{\mergeduparrowlen}}$%
        }%
      }%
    }%
  }%
  \makebox[0pt][l]{%
    \kern.5\wd0\kern.8pt
    \makebox[0pt][c]{%
      \raisebox{-\dp0}[0pt][0pt]{%
        \rotatebox{-90}{%
          $\xrightarrow{\hspace{\mergeddownarrowlen}}$%
        }%
      }%
    }%
  }%
  \box0
  \endgroup
}
\makeatletter
\newcommand{\mergedrowband}[1]{%
  \multicolumn{2}{@{}c@{}}{} &
  \multicolumn{9}{@{}l@{}}{%
    \color{#1}%
    \leaders\hrule
      height\ht\@arstrutbox
      depth\dp\@arstrutbox
      \hfill\kern0pt}%
  \\[-\dimexpr\ht\@arstrutbox+\dp\@arstrutbox\relax]}
\makeatother

\newcommand{\oodQwenPTrueAUROC}{\mergedms{.555}{.010}}
\newcommand{\oodQwenSelfProbingAUROC}{\mergedms{.508}{.028}}
\newcommand{\oodQwenPromptEnsembleAUROC}{\mergedms{.519}{.019}}
\newcommand{\oodQwenSAPLMAAUROC}{\mergedbest{.697}{.006}}
\newcommand{\oodMedPTrueAUROC}{\mergedms{.594}{.015}}
\newcommand{\oodMedSelfProbingAUROC}{\mergedms{.527}{.045}}
\newcommand{\oodMedPromptEnsembleAUROC}{\mergedms{.537}{.031}}
\newcommand{\oodMedSAPLMAAUROC}{\mergedbest{.687}{.018}}
\begingroup
\small
\setlength{\tabcolsep}{.4pt}
\renewcommand{\arraystretch}{1.08}
\begin{tabular*}{\textwidth}{@{\extracolsep{\fill}}cll*{4}{c}@{\hspace{1.5pt}}*{4}{c}@{}}
\toprule
\multirow{2}{*}{\shortstack{}} &
\multirow{2}{*}{\shortstack{\textbf{Confidence}\\\textbf{proxy}}} &
\multirow{2}{*}{\textbf{Baseline}} &
\multicolumn{4}{c}{\textbf{SLAKE-test} ($n{=}1061$)} &
\multicolumn{4}{c}{\textbf{O.O.D. Averaged} ($n{=}4584$)} \\
\cmidrule(lr){4-7}\cmidrule(lr){8-11}
& & &
\textbf{Acc.} &
\textbf{AUROC} &
\shortstack{\textbf{Natural}\\\textbf{CCG AUC}} &
\shortstack{\textbf{Strict}\\\textbf{CCG AUC}} &
\textbf{Acc.} &
\textbf{AUROC} &
\shortstack{\textbf{Natural}\\\textbf{CCG AUC}} &
\shortstack{\textbf{Strict}\\\textbf{CCG AUC}} \\
\midrule

\mergedrowband{MergedBaseGray}
&
\multirow{5}{*}{\shortstack{Verbalized\\token}} &
Base & \ensuremath{.443} & \ensuremath{.579} & \ensuremath{.587} & \ensuremath{.423} &
\ensuremath{.450} & \ensuremath{.696} &
\ensuremath{\mathbf{.782}} & \ensuremath{\mathbf{.622}} \\
& & RLCR&
\mergedms{.608}{.009} & \mergedms{.822}{.014} & \mergedms{.533}{.031} & \mergedms{.514}{.036} &
\mergedms{.492}{.012} & \mergedms{.613}{.008} & \mergedms{.516}{.022} & \mergedms{.509}{.028} \\
& & DCPO &
\mergedms{.640}{.003} & \mergedms{.789}{.007} & \mergedms{.554}{.021} & \mergedms{.540}{.022} &
\mergedms{.491}{.004} & \mergedms{.622}{.005} & \mergedms{.505}{.018} & \mergedms{.509}{.011} \\
& & \shortstack{VL-Cal.} &
\mergedms{.591}{.010} & \mergedms{.756}{.026} & \mergedms{.553}{.016} & \mergedms{.500}{.034} &
\mergedms{.497}{.010} & \mergedms{.633}{.020} & \mergedms{.566}{.045} & \mergedms{.559}{.042} \\
\cmidrule(lr){4-4}\cmidrule(lr){8-8}
& & \shortstack{GRPO(RLVR)} &
& \mergedms{.535}{.059} & \mergedms{.516}{.031} & \mergedms{.507}{.018} &
& \mergedms{.507}{.008} & \mergedms{.529}{.051} & \mergedms{.516}{.027} \\
& \multirow{3}{*}{\shortstack{Prompted\\self-check}} & P(True) &
& \mergedms{.609}{.012} & \mergedms{.580}{.038} & \mergedms{.533}{.052} &
& \oodQwenPTrueAUROC & \mergedms{.514}{.101} & \mergedms{.412}{.101} \\
& & \shortstack{Self-prob.} &
& \mergedms{.517}{.033} & \mergedms{.532}{.050} & \mergedms{.518}{.032} &
& \oodQwenSelfProbingAUROC & \mergedms{.491}{.056} & \mergedms{.442}{.103} \\
& & \shortstack{Prompt ensemble} &
& \mergedms{.534}{.034} & \mergedms{.524}{.078} & \mergedms{.488}{.051} &
& \oodQwenPromptEnsembleAUROC & \mergedms{.518}{.057} & \mergedms{.453}{.072} \\
& & SAPLMA &
& \mergedms{.942}{.004} & \mergedms{.715}{.044} & \mergedms{.693}{.068} &
& \oodQwenSAPLMAAUROC & \mergedms{.550}{.145} & \mergedms{.484}{.199} \\
& & BICR &
& \mergedms{.934}{.002} & \mergedms{.723}{.039} & \mergedms{.705}{.050} &
& \mergedms{.644}{.005} & \mergedms{.530}{.026} & \mergedms{.597}{.029} \\
\mergedrowband{MergedOursBlue}
\multirow{-11}{*}{\rotatebox[origin=c]{90}{Qwen3VL-2B}} &
\multirow{-3}{*}{\shortstack{Hidden\\state}} & \textbf{DualRead} &
\multirow{-7}{*}[11pt]{\mergedsharedacc{\mergedms{.639}{.009}}} &
\mergedbest{.952}{.004} &
\mergedbest{.784}{.009} &
\mergedbest{.738}{.039} &
\multirow{-7}{*}[11pt]{\mergedsharedacc{\mergedms{.494}{.004}}} &
\mergedms{.666}{.021} & \mergedms{.624}{.067} &
\mergedms{.561}{.099} \\
\midrule

\mergedrowband{MergedBaseGray}
&
\multirow{5}{*}{\shortstack{Verbalized\\token}} &
Base & \ensuremath{.478} & \ensuremath{.615} & \ensuremath{.535} & \ensuremath{.542}  &
\ensuremath{.494} & \ensuremath{.590} & \ensuremath{.502} &
\ensuremath{.375} \\
& & RLCR &
\mergedms{.657}{.027} & \mergedms{.771}{.021} & \mergedms{.527}{.020} & \mergedms{.507}{.020} &
\mergedms{.576}{.001} & \mergedms{.601}{.025} & \mergedms{.519}{.004} & \mergedms{.495}{.008} \\
& & DCPO &
\mergedms{.678}{.003} & \mergedms{.752}{.018} & \mergedms{.509}{.025} & \mergedms{.519}{.033} &
\mergedms{.561}{.021} & \mergedms{.621}{.013} & \mergedms{.568}{.012} & \mergedms{.557}{.016} \\
& & \shortstack{VL-Cal.} &
\mergedms{.587}{.005} & \mergedms{.851}{.005} & \mergedms{.608}{.039} & \mergedms{.580}{.019} &
\mergedms{.560}{.005} & \mergedms{.661}{.025} & \mergedms{.598}{.043} & \mergedms{.554}{.027} \\
\cmidrule(lr){4-4}\cmidrule(lr){8-8}
& & \shortstack{GRPO(RLVR)} &
& \mergedms{.506}{.011} & \mergedms{.503}{.005} & \mergedms{.504}{.007} &
& \mergedms{.502}{.003} & \mergedms{.504}{.006} & \mergedms{.506}{.011} \\
& \multirow{3}{*}{\shortstack{Prompted\\self-check}} & P(True) &
& \mergedms{.574}{.026} & \mergedms{.634}{.054} & \mergedms{.678}{.060} &
& \oodMedPTrueAUROC & \mergedms{.538}{.082} & \mergedms{.475}{.161} \\
& & \shortstack{Self-prob.} &
& \mergedms{.597}{.110} & \mergedms{.557}{.050} & \mergedms{.561}{.046} &
& \oodMedSelfProbingAUROC & \mergedms{.558}{.055} & \mergedms{.571}{.061} \\
& & \shortstack{Prompt ensemble} &
& \mergedms{.626}{.073} & \mergedms{.548}{.070} & \mergedms{.550}{.099} &
& \oodMedPromptEnsembleAUROC & \mergedms{.539}{.075} & \mergedms{.512}{.097} \\
& & SAPLMA &
& \mergedms{.942}{.011} & \mergedms{.814}{.063} & \mergedms{.767}{.080} &
& \oodMedSAPLMAAUROC & \mergedms{.550}{.143} & \mergedms{.636}{.174} \\
& & BICR &
& \mergedms{.930}{.007} & \mergedms{.801}{.023} &
\mergedbest{.804}{.040} &
& \mergedms{.635}{.022} & \mergedms{.630}{.053} & \mergedms{.644}{.108} \\
\mergedrowband{MergedOursBlue}
\multirow{-11}{*}{\rotatebox[origin=c]{90}{MedGemma1.5-4B}} &
\multirow{-3}{*}{\shortstack{Hidden\\state}} & \textbf{DualRead} &
\multirow{-7}{*}[11pt]{\mergedsharedacc{\mergedms{.677}{.011}}} &
\mergedbest{.947}{.005} &
\mergedbest{.819}{.030} & \mergedms{.786}{.033} &
\multirow{-7}{*}[11pt]{\mergedsharedacc{\mergedms{.586}{.005}}} &
\mergedms{.647}{.034} &
\mergedbest{.634}{.162} &
\mergedbest{.683}{.088} \\
\bottomrule
\end{tabular*}
\endgroup
\end{table*}
\subsection{Datasets and Evaluation Metrics}

We evaluate all methods on SLAKE-test~\citep{liu2021slake} with $1{,}061$ questions as the in-distribution benchmark. To assess out-of-distribution generalization, we use six medical VQA datasets: VQA-RAD~\citep{lau2018vqarad} with $600$ questions, PathVQA~\citep{he2020pathvqa} with $1{,}000$ questions, PMC-VQA~\citep{zhang2024pmcvqa} with $1{,}000$ questions, OmniMedVQA~\citep{hu2024omnimedvqa} with $920$ questions, ReXVQA~\citep{pal2026rexvqa} with $1{,}000$ questions, and the closed-answer subset of VQA-Med 2019~\citep{benabacha2019vqamed} with $64$ questions, yielding $4{,}584$ OOD examples in total.  No OOD examples are used to fit the actors, confidence readers, or calibration temperatures. For the OOD summary, each metric is first computed independently on each domain and then macro-averaged across the six domains, preventing datasets with more examples from dominating the result. We report answer accuracy (Acc.) and the area under the receiver operating characteristic curve (AUROC) for correctness discrimination. Natural CCG-AUC and Strict CCG-AUC are also reported. The ablation and cost analyses add Brier and the tie-aware
area under the risk--coverage curve (AURC). All post-hoc confidence proxy use the same frozen RLVR actor and therefore share its answer accuracy while differing only in their confidence estimates. Please refer to the supplementary material for other detailed calibration metrics including ECE and Brier. 

\subsection{Implementation Details}
We instantiated the capability actor using Qwen3-VL-2B-Instruct~\cite{qwen3technicalreport} and MedGemma1.5-4B~\cite{sellergren2026medgemma} and fully post-trained each backbone on SLAKE-train with GRPO. GRPO used a group size of $8$, the AdamW optimizer with a learning rate of $1 \times 10^{-6}$, a global batch size of $128$, and three passes over the training set. Except for the pretrained Base models, all methods were evaluated over three random seeds. For a fair comparison, all policy baselines shared the same training data, prompts, grader, and optimization protocol, while all hidden-state baselines used the same frozen RLVR actor and its greedily generated answers. For the sampling-based baselines, SC@$K$ measures agreement among $K$ stochastic completions, while semantic uncertainty clusters the same eight completions by canonical answer.

DualRead extracted the pre-answer and post-answer representations through one teacher-forced replay,   with the cosine-similarity pooling temperature fixed to $\kappa=\textnormal{0.1}$ throughout all experiments. The post-answer linear heads were fitted sequentially for $1{,}000$ steps using binary cross-entropy with a learning rate of $1 \times 10^{-3}$, while the PSR head used an MLP with two hidden layers of $256$ and $128$ units and early stopping. Reader fitting and temperature scaling used image-disjoint partitions of SLAKE-train without target-domain adaptation. The calibrated branches were fused using a fixed equal-weight mean in log-odds space. All experiments were performed on a single node with 8 NVIDIA H200 GPUs (141 GB each) under Ubuntu 22.04.5 LTS, using PyTorch 2.8.0 with CUDA 12.8.

\subsection{Results}
Table~\ref{tab:main} compares DualRead with three families of confidence
proxy on Qwen3VL-2B and MedGemma1.5-4B of verbalized-token methods,
prompted self-checking methods, and hidden-state probes. Base, RLCR, DCPO,
and VL-Calibration use different actor outputs. In contrast, GRPO (RLVR),
P(True), self-probing, prompt ensemble, SAPLMA, BICR, and DualRead are
evaluated on exactly the same frozen RLVR actor outputs and therefore share
the same accuracy. This design isolates confidence estimation from changes
in answer generation.

Although RLVR achieves competitive answer accuracy, its
verbalized confidence remains weakly aligned with correctness. Prompted
self-checking provides only limited improvements, whereas hidden-state methods
perform substantially better. DualRead achieves the best ID AUROC and Natural
CCG AUC on both backbones, as well as the best Strict CCG AUC on Qwen3VL-2B,
demonstrating that its confidence is both reliable and grounded in the relevant
visual evidence.

The OOD results further demonstrate the generalizability of DualRead. Without
changing the actor predictions, DualRead consistently improves AUROC and both
CCG metrics over the confidence verbalized by RLVR. It achieves the strongest
OOD grounding performance on MedGemma1.5-4B and remains competitive on
Qwen3VL-2B. SAPLMA and BICR perform better on several individual OOD metrics,
suggesting that cross-domain confidence estimation remains dependent on the
backbone and confidence proxy. Overall, DualRead provides the most consistent
improvement in confidence reliability and grounding while fully preserving the
accuracy of the underlying RLVR actor.

\subsubsection{Ablation study.}
\begin{table}[!t]
\centering
\caption{Component ablation of DualRead. PER combines P-State and P-Visual;
DualRead further combines PER with PSR. CCG-AUC here is the Natural variant on SLAKE-test.}
\label{tab:cascade}
\footnotesize
\newcommand{\cascadems}[2]{\ensuremath{#1_{\pm\!#2}}}
\newcommand{\cascadebest}[2]{\ensuremath{\mathbf{#1}_{\pm\!#2}}}
\setlength{\tabcolsep}{0pt}
\renewcommand{\arraystretch}{1.06}
\begin{tabular*}{\columnwidth}{@{\extracolsep{\fill}}lcccccc@{}}
\toprule
\multirow{2}{*}{\textbf{Variant}} &
\multicolumn{3}{c}{\textbf{Components}} &
\multirow{2}{*}{\textbf{AUROC}$\uparrow$} &
\multirow{2}{*}{\textbf{Brier}$\downarrow$} &
\multirow{2}{*}{\textbf{CCG-AUC}$\uparrow$}\\
\cmidrule(lr){2-4}
& \textbf{PSR} & \textbf{State} & \textbf{Visual} & & & \\
\midrule
\multicolumn{7}{@{}l}{\emph{Qwen3VL-2B}}\\
PSR               & $\checkmark$ & $\times$     & $\times$     & \cascadems{.926}{.008} & \cascadems{.101}{.005} & \cascadems{.733}{.044}\\
P-State           & $\times$     & $\checkmark$ & $\times$     & \cascadems{.928}{.006} & \cascadems{.099}{.007} & \cascadems{.686}{.027}\\
PER               & $\times$     & $\checkmark$ & $\checkmark$ & \cascadems{.947}{.003} & \cascadems{.084}{.001} & \cascadems{.760}{.038}\\
\textbf{DualRead} & $\checkmark$ & $\checkmark$ & $\checkmark$ & \cascadebest{.952}{.004} & \cascadebest{.081}{.004} & \cascadebest{.784}{.009}\\
\addlinespace[2pt]
\multicolumn{7}{@{}l}{\emph{MedGemma1.5-4B}}\\
PSR               & $\checkmark$ & $\times$     & $\times$     & \cascadems{.914}{.009} & \cascadems{.105}{.007} & \cascadems{.725}{.053}\\
P-State           & $\times$     & $\checkmark$ & $\times$     & \cascadems{.934}{.002} & \cascadems{.086}{.003} & \cascadems{.769}{.050}\\
PER               & $\times$     & $\checkmark$ & $\checkmark$ & \cascadems{.945}{.006} & \cascadems{.079}{.005} & \cascadebest{.825}{.040}\\
\textbf{DualRead} & $\checkmark$ & $\checkmark$ & $\checkmark$ & \cascadebest{.947}{.005} & \cascadebest{.076}{.004} & \cascadems{.819}{.030}\\
\bottomrule
\end{tabular*}
\end{table}

To assess the contribution of each component, we conduct an ablation study,
as shown in Table~\ref{tab:cascade}. Both PSR and P-State provide competitive
confidence estimates, indicating that useful correctness information exists
both before and after answer generation. Incorporating the visual residual
into P-State consistently improves discrimination, calibration, and
confidence grounding, demonstrating the importance of answer-conditioned
visual evidence. Combining PSR and PER in DualRead further achieves the
highest AUROC and lowest Brier score on both backbones, together with the
best CCG-AUC on Qwen3VL-2B. Although PER slightly outperforms DualRead in
CCG-AUC on MedGemma1.5-4B, DualRead remains superior in overall reliability.
These results demonstrate that the pre-answer and post-answer branches capture
complementary confidence information.

\subsection{Deeper Analysis and Discussions}
\begin{figure}[tb]
\centering
\includegraphics[width=\linewidth]{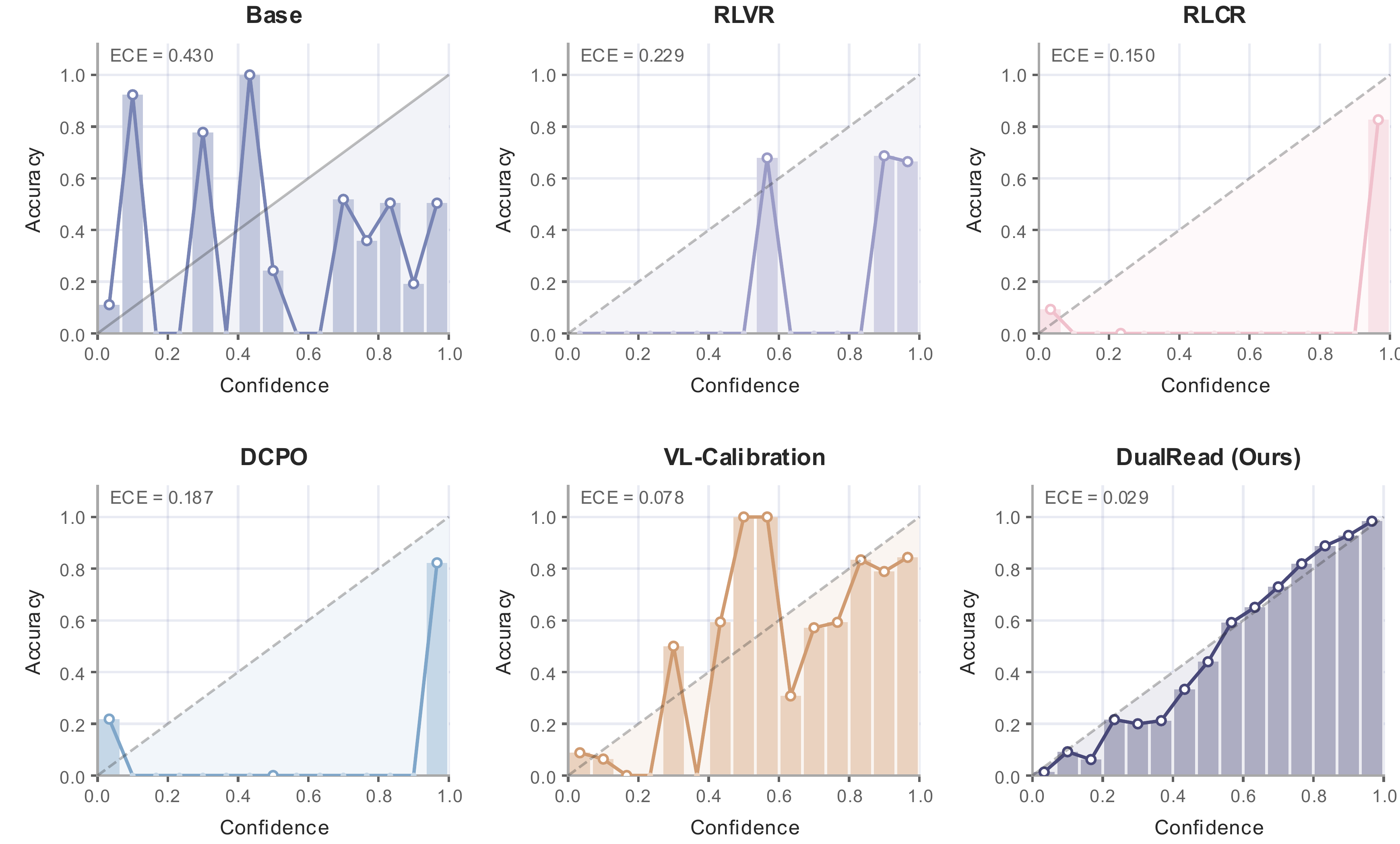}
\caption{Reliability diagrams on SLAKE-test. The dashed diagonal denotes perfect calibration.}
\label{fig:reliability}
\end{figure}

\begin{figure*}[!t]
\centering
\includegraphics[width=\linewidth]{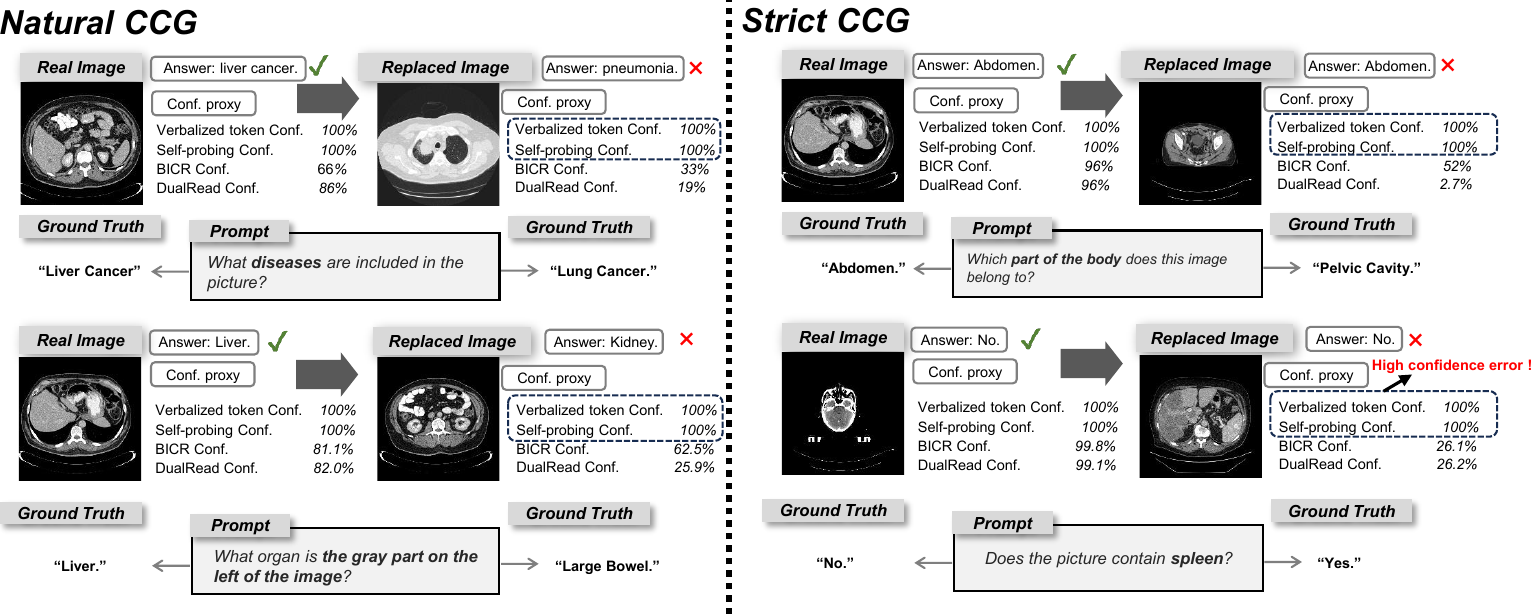}
\caption{Qualitative examples of counterfactual confidence grounding.
Natural CCG allows the answer to change after same-question image replacement, whereas Strict CCG retains the answer while flipping its correctness.}
\label{fig:grounding}
\end{figure*}
\subsubsection{Policy optimization compresses verbalized confidence.}
Figure~\ref{fig:reliability} reveals a distinction that is not fully captured by aggregate metrics. The confidence scores produced by the Base model are broadly distributed but poorly calibrated, whereas RLVR concentrates most predictions near the maximum-confidence endpoint. Calibration-aware policy optimization partially changes this distribution, but its verbalized scores remain highly concentrated or nearly binary, limiting their ability to rank intermediate-risk cases. In contrast, DualRead produces a continuous confidence distribution that more closely follows the empirical accuracy curve. This result suggests that RLVR does not necessarily remove correctness-related information from the model; rather, policy optimization compresses the interface through which this information is verbalized. Reading confidence from the frozen internal states therefore recovers useful resolution without modifying the actor or its answers.

\begin{table}[!t]
\centering
\caption{Compute--reliability trade-off. $t$ is the required greedy
generation, $\rho$ one prompt-only view, and $\tau$ DualRead's natural-trajectory replay. Lower AURC means better selective-prediction risk at matched coverage.}
\label{tab:sc_tradeoff}
\small
\setlength{\tabcolsep}{4pt}
\begin{tabular}{@{}llccc@{}}
\toprule
Methods & Cost & AUROC$\uparrow$ & Brier$\downarrow$ & AURC$\downarrow$\\
\midrule
\multicolumn{5}{@{}l}{\emph{Qwen3VL-2B, ID: SLAKE-test}}\\
SC@2 & $2t$ & $0.742_{\pm.006}$ & $0.165_{\pm.002}$ & $0.234_{\pm.005}$\\
SC@8 & $8t$ & $0.831_{\pm.003}$ & $0.140_{\pm.003}$ & $0.179_{\pm.006}$\\
Semantic unc. & $8t$ & $0.831_{\pm.002}$ & $0.135_{\pm.001}$ & $0.179_{\pm.007}$\\
BICR & $t{+}2\rho$ & $0.934_{\pm.002}$ & $0.099_{\pm.003}$ & $0.101_{\pm.004}$\\
\textbf{DualRead} & $t{+}\tau$ & $\mathbf{0.952}_{\pm.004}$ & $\mathbf{0.083}_{\pm.002}$ & $\mathbf{0.094}_{\pm.003}$\\
\addlinespace[2pt]
\multicolumn{5}{@{}l}{\emph{Qwen3VL-2B, OOD: VQA-RAD}}\\
SC@2 & $2t$ & $0.699_{\pm.011}$ & $0.226_{\pm.007}$ & $0.428_{\pm.009}$\\
SC@8 & $8t$ & $0.765_{\pm.016}$ & $0.207_{\pm.010}$ & $\mathbf{0.379}_{\pm.007}$\\
Semantic unc. & $8t$ & $\mathbf{0.766}_{\pm.017}$ & $\mathbf{0.205}_{\pm.011}$ & $0.380_{\pm.009}$\\
BICR & $t{+}2\rho$ & $0.686_{\pm.019}$ & $0.266_{\pm.020}$ & $0.413_{\pm.011}$\\
\textbf{DualRead} & $t{+}\tau$ & $0.728_{\pm.010}$ & $0.234_{\pm.006}$ & $0.387_{\pm.009}$\\
\midrule
\multicolumn{5}{@{}l}{\emph{MedGemma1.5-4B, ID: SLAKE-test}}\\
SC@2 & $2t$ & $0.659_{\pm.034}$ & $0.181_{\pm.010}$ & $0.248_{\pm.007}$\\
SC@8 & $8t$ & $0.740_{\pm.040}$ & $0.165_{\pm.013}$ & $0.205_{\pm.012}$\\
Semantic unc. & $8t$ & $0.740_{\pm.040}$ & $0.161_{\pm.014}$ & $0.205_{\pm.012}$\\
BICR & $t{+}2\rho$ & $0.930_{\pm.007}$ & $0.100_{\pm.007}$ & $0.084_{\pm.006}$\\
\textbf{DualRead} & $t{+}\tau$ & $\mathbf{0.947}_{\pm.005}$ & $\mathbf{0.082}_{\pm.002}$ & $\mathbf{0.079}_{\pm.005}$\\
\addlinespace[2pt]
\multicolumn{5}{@{}l}{\emph{MedGemma1.5-4B, OOD: VQA-RAD}}\\
SC@2 & $2t$ & $0.678_{\pm.011}$ & $0.232_{\pm.007}$ & $0.399_{\pm.019}$\\
SC@8 & $8t$ & $0.739_{\pm.009}$ & $0.218_{\pm.005}$ & $0.354_{\pm.010}$\\
Semantic unc. & $8t$ & $0.746_{\pm.005}$ & $\mathbf{0.209}_{\pm.003}$ & $0.352_{\pm.010}$\\
BICR & $t{+}2\rho$ & $0.723_{\pm.002}$ & $0.288_{\pm.023}$ & $0.358_{\pm.008}$\\
\textbf{DualRead} & $t{+}\tau$ & $\mathbf{0.756}_{\pm.016}$ & $0.217_{\pm.017}$ & $\mathbf{0.331}_{\pm.006}$\\
\bottomrule
\end{tabular}

\end{table}

\subsubsection{DualRead responds to visual evidence.}
The counterfactual examples in Figure~\ref{fig:grounding} illustrate the difference between verbalized confidence and hidden-state confidence. When the original image is replaced by a same-question hard negative, the actor may retain the same answer and report the same near-maximum verbalized confidence even though the answer becomes incorrect. DualRead instead assigns substantially lower confidence to the replaced image. Because the question and canonical answer remain unchanged, the confidence difference cannot be explained solely by question difficulty or answer wording. These examples provide an intuitive view of the behavior measured by Strict CCG-AUC: confidence decreases when the visual evidence supporting an otherwise identical answer is removed. The examples are illustrative rather than standalone evidence and should be interpreted together with the aggregate CCG-AUC results.

\subsubsection{DualRead provides an efficient compute--reliability trade-off.}
Table~\ref{tab:sc_tradeoff} further
compares DualRead with repeated-sampling and counterfactual-view
baselines under their corresponding inference costs. DualRead achieves
a favorable reliability--compute trade-off on the in-domain setting
without requiring multiple full generations. Under distribution shift,
repeated sampling remains competitive on some metrics, whereas DualRead
provides consistently strong reliability with substantially lower
additional inference. These results indicate that the benefit of
DualRead arises from extracting complementary evidence along the
natural trajectory rather than merely increasing the number of model
queries.

\section{Conclusion}
\label{sec:conclusion}

Reliable confidence can be obtained from hidden state, rather than generated by a GRPO-trained medical
VLM. DualRead freezes the actor and fuses PSR with PER, improving discrimination,
calibration, and selective prediction without changing its answers. Under same-question
hard negatives, the resulting confidence responds to decisive visual evidence while the
same actor's verbalized token does not, consistently across both backbones. Our evidence is limited to short-answer medical VQA on $2$--$4$B models; larger models,
report generation, and non-uniform open/closed behavior remain open. Future work should
test state-based training signals without recreating token-interface collapse and extend
the counterfactual protocol to entity-level claims in clinical reports.

\FloatBarrier
\bibliographystyle{aaai2027}
\bibliography{reference}
\clearpage
\section*{Technical Supplement}
\appendix
\setcounter{secnumdepth}{1}
\titleformat{\section}{\centering\large\bfseries}{\thesection.}{.4em}{}
%

This supplement documents the frozen protocol, the exact configurations, and the extended measurements behind the main paper.

\section{GRPO Formulation}
\label{app:grpo}

For each input $x_i=(I_i,q_i)$, the actor $\pi_\theta$ samples a group of $G$
responses $\{y_{i,g}\}_{g=1}^{G}$.  Each response receives a verifiable task
reward
\begin{equation}
R_{i,g}=z_{i,g}+R_{\mathrm{fmt},i,g},
\end{equation}
where $z_{i,g}\in\{0,1\}$ denotes answer correctness and
$R_{\mathrm{fmt},i,g}$ penalizes malformed responses.  GRPO converts these
rewards into group-relative advantages,
\begin{equation}
A_{i,g}
=
\frac{R_{i,g}-\operatorname{mean}_{h}(R_{i,h})}
{\operatorname{std}_{h}(R_{i,h})+\epsilon},
\end{equation}
and applies a clipped policy objective to increase the likelihood of responses
that outperform others sampled for the same input, yielding the trained actor
$\theta^{*}=\arg\max_{\theta}\mathcal{J}_{\mathrm{GRPO}}(\theta;R)$.

\section{Hyperparameters}
\label{app:hparams}

Table~\ref{tab:hparams} lists the complete frozen configuration for GRPO actor
training, DualRead head fitting, and evaluation.

\begin{table*}[!t]
\centering
\small
\setlength{\tabcolsep}{6pt}
\renewcommand{\arraystretch}{1.05}
\begin{tabular}{@{}lp{0.67\textwidth}@{}}
\toprule
\multicolumn{2}{@{}l}{\textit{GRPO actor training (shared; reward differs per method)}}\\
\midrule
Advantage estimator            & GRPO, group size $G=8$ \\
Rollout temperature / top-$p$  & $1.0$ / $1.0$ \\
PPO clip (low / high)          & $0.2$ / $0.2$ \\
KL penalty                     & disabled ($\mathrm{coef}=0$) \\
Optimizer                      & AdamW, lr $1\!\times\!10^{-6}$, $\beta=(0.9,0.95)$ \\
LR schedule / warmup           & constant, $5\%$ warmup \\
Weight decay / grad-norm clip  & $1\!\times\!10^{-2}$ / $1.0$ \\
Precision / attention          & BF16 / SDPA \\
Trainable params               & full model incl.\ vision tower \\
Prompt batch / global batch    & $128$ / $128$ \\
Max prompt / completion length & $4096$ / $256$ (Qwen)$^{\dagger}$ \\
Training exposure              & $3$ passes over SLAKE-train ($117$ steps) \\
Seeds                          & $\{17, 29, 43\}$ \\
Hardware                       & one node with $8{\times}$H200 (141\,GB); each run uses
                                 $4$ GPUs (Qwen) or $2$ GPUs (MedGemma) \\
Software                       & Ubuntu 22.04.5 LTS, PyTorch 2.8.0, CUDA 12.8 \\
\midrule
\multicolumn{2}{@{}l}{\textit{Hidden-state head fitting (DualRead)}}\\
\midrule
PSR head (pre-answer)          & MLP $256$--$128$, BCE, AdamW lr $1.276\!\times\!10^{-4}$,
                                 weight decay $2.017\!\times\!10^{-6}$, batch $32$,
                                 $\le200$ epochs, early stopping (patience $20$);
                                 architecture frozen \\
PER heads (post-answer)        & P-State then P-Visual, linear pointwise BCE,
                                 $1000$ full-batch steps, lr $1\!\times\!10^{-3}$ \\
$L2$ (P-State / P-Visual)      & $1\!\times\!10^{-4}$ / $1\!\times\!10^{-3}$ \\
Visual pooling                 & cosine-similarity pooling, temperature $\kappa=0.1$ \\
Fusion                         & log-odds mean of calibrated PSR and PER,
                                 weight $0.5$ (frozen a priori) \\
Calibration                    & one scalar temperature per branch, fit on
                                 held-out SLAKE-train images \\
\midrule
\multicolumn{2}{@{}l}{\textit{Evaluation}}\\
\midrule
Decoding                       & greedy (temperature $0$) \\
Invalid-confidence policy      & retain, impute $c=0.5$, report parse rate \\
Bootstrap                      & paired image-cluster, $B=10{,}000$, $95\%$ percentile \\
\bottomrule
\end{tabular}
\caption{Frozen hyperparameters. $^{\dagger}$MedGemma uses a 512-token prompt limit without SLAKE truncation.}
\label{tab:hparams}
\end{table*}

\section{Dataset Details}
\label{app:datasets}

Table~\ref{tab:dataset_audit} gives the exact partitions consumed by the
experiments: SLAKE, VQA-RAD, PathVQA, PMC-VQA, OmniMedVQA, ReXVQA, and
VQA-Med 2019.  Every split is represented by a
frozen manifest with one row per question and an image identifier used for
disjoint fitting, subgroup analysis, and clustered resampling.

\begin{table*}[!t]
\centering
\small
\renewcommand{\arraystretch}{1.07}
\begin{tabular}{@{}llrrrr@{}}
\toprule
Dataset & Role & QA & Images & Open & Closed\\
\midrule
SLAKE-train & actor/head fit & 4919 & 450 & 2976 & 1943\\
SLAKE-val. & selection only & 1053 & 96 & 631 & 422\\
SLAKE-test & ID report & 1061 & 96 & 645 & 416\\
VQA-RAD & OOD report & 600 & 270 & 236 & 364\\
PathVQA & OOD report & 1000 & 706 & 500 & 500\\
PMC-VQA & OOD report & 1000 & 1000 & 994 & 6\\
OmniMedVQA & OOD report & 920 & 920 & 0 & 920\\
ReXVQA & OOD report & 1000 & 1000 & 0 & 1000\\
VQA-Med 2019 (closed) & OOD report & 64 & 64 & 0 & 64\\
\bottomrule
\end{tabular}
\caption{Frozen dataset inventory; images are unique identifiers for leakage control and clustered resampling.}
\label{tab:dataset_audit}
\end{table*}

\paragraph{SLAKE.}
Each question carries answer\_type in $\{$open, closed$\}$ and a content
family (\emph{modality, organ, plane, position, abnormality, size, shape, color,
quantity, knowledge}).  Metrics are recomputed from item-level predictions
inside each subgroup rather than derived from the overall value.  SLAKE
validation is used only for checkpoint/model selection and deployable threshold
selection; it never enters a test metric.

\paragraph{VQA-RAD.}
The frozen 600-question manifest contains 270 images and preserves the short
free-form answer contract used by SLAKE.  It is used as a zero-touch transport
test.

\paragraph{PathVQA.}
The complete PathVQA test manifest contains 6719 questions.  Before model
evaluation we drew an answer-type-balanced subset: 500 open
and 500 closed questions, at most two questions per image, uniform sampling
within answer type, with the corpus mixture otherwise preserved.  The resulting
set contains 706 unique pathology images and retains the same free-form
short-answer output contract as SLAKE/VQA-RAD.  Its content-family counts are
10 modality, 51 position, 429 presence, 4 size, and 506 other questions.  The
selection rule, IDs, and images are fixed across all methods and seeds.

\paragraph{Option-visible OOD datasets.}
PMC-VQA, OmniMedVQA, and ReXVQA use the semantic-answer multiple-choice
interface: the actor sees all four options and
places the complete selected option text in the answer span. The respective
frozen factual manifests contain 1000, 920, and 1000 questions. They are part
of the six-domain OOD average in the main paper; shuffled-image variants remain
supplementary controls only. Per-domain results are given in
Tables~\ref{tab:ood_acc} and~\ref{tab:calib}.

\paragraph{VQA-Med 2019 closed subset.}
The de-duplicated VQA-Med 2019 test view contains 496 questions after removing four images duplicated in VQA-RAD. We uses only its frozen 64-question closed subset, selected by question identity before scoring for every method and seed.

\section{Evaluation Details}
\label{app:evaluation_protocol}

\paragraph{One deterministic correctness grader.}
Training rewards and all reported labels use the same non-LLM grader.  It
lowercases, trims punctuation and whitespace, drops one leading article, applies a guarded singular/plural fold, and maps a small frozen set of equivalences (e.g., \emph{xray}/\emph{radiograph}, \emph{CT}/\emph{computed tomography}, and yes/no variants). Laterality is never collapsed. A prediction is correct iff its non-empty canonical form exactly matches the canonical gold answer. Empty, duplicated, or unparseable answers score zero. No synonym was added after
test/OOD inspection and no LLM judge is used for headline results.

\paragraph{Output and confidence parsing.}
The shared contract is
<reasoning>...</reasoning>\allowbreak
<answer>...</answer>\allowbreak
<confidence>NN</confidence>,
where NN is one integer in $[0,100]$.  Exactly one non-empty answer
block and one integer confidence block are required.  Floats, percentages,
out-of-range values, missing tags, or duplicated tags make confidence invalid.
Invalid rows are \emph{retained} and assigned $c=0.5$ for evaluation; the valid
fraction is reported in Table~\ref{tab:channel}.  VL-Calibration retains its native vision/reasoning confidence blocks in $[0,10]$ and uses the method's fixed harmonic aggregation.  Base uses normalized logits over the elicited confidence vocabulary, so token parse rate is not applicable. State readers always return a finite scalar and consequently have parse rate one.

\begin{table*}[!t]
\centering
\footnotesize
\begin{minipage}[t]{.47\textwidth}
\centering
\textit{(a) Frozen data inventory}\par\smallskip
\setlength{\tabcolsep}{2pt}
\renewcommand{\arraystretch}{1.04}
\begin{tabular*}{\linewidth}{@{\extracolsep{\fill}}llrrrr@{}}
\toprule
Dataset & Role & QA & Images & Open & Closed\\
\midrule
SLAKE-train & fit & 4919 & 450 & 2976 & 1943\\
SLAKE-val. & select & 1053 & 96 & 631 & 422\\
SLAKE-test & ID & 1061 & 96 & 645 & 416\\
VQA-RAD & OOD & 600 & 270 & 236 & 364\\
PathVQA & OOD & 1000 & 706 & 500 & 500\\
PMC-VQA & OOD & 1000 & 1000 & 994 & 6\\
OmniMedVQA & OOD & 920 & 920 & 0 & 920\\
ReXVQA & OOD & 1000 & 1000 & 0 & 1000\\
VQA-Med19-C & OOD & 64 & 64 & 0 & 64\\
\bottomrule
\end{tabular*}
\end{minipage}\hfill
\begin{minipage}[t]{.51\textwidth}
\centering
\textit{(b) Confidence-channel audit}\par\smallskip
\setlength{\tabcolsep}{2pt}
\renewcommand{\arraystretch}{1.04}
\begin{tabular*}{\linewidth}{@{\extracolsep{\fill}}llrrrr@{}}
\toprule
& & \multicolumn{2}{c}{Parse rate} & \multicolumn{2}{c}{Support}\\
\cmidrule(lr){3-4}\cmidrule(lr){5-6}
Backbone & Interface & SLAKE & VQA-RAD & Unique & Endpt.\\
\midrule
\multirow{3}{*}{Qwen}
 & RLVR token & $1.000  $ & $1.000  $ & $2$ & $1.000$\\
 & RLCR token & $1.000  $ & $1.000  $ & $2$ & $1.000$\\
 & BICR / DualRead & $1.000  $ & $1.000  $ & $1061$ & $0.000$\\
\midrule
\multirow{3}{*}{MedGemma}
 & RLVR token & $1.000  $ & $1.000  $ & $1$ & $1.000$\\
 & RLCR token & $1.000  $ & $1.000  $ & $2$ & $1.000$\\
 & BICR / DualRead & $1.000  $ & $1.000  $ & $1061$ & $0.000$\\
\bottomrule
\end{tabular*}
\end{minipage}
\caption{Frozen data and confidence-channel audits; invalid confidence is retained at $c=0.5$.}
\label{tab:data}\label{tab:channel}
\end{table*}

\paragraph{Metric definitions.}
For confidence $c_i\in[0,1]$ and correctness $z_i\in\{0,1\}$, Accuracy is
$n^{-1}\sum_i z_i$ and Brier is
$n^{-1}\sum_i(c_i-z_i)^2$.  AUROC is the probability that a randomly selected correct item has higher confidence than a randomly selected incorrect item, with a tie contributing $0.5$.  ECE-15 uses the fixed intervals $[0,\frac1{15}),\ldots,[\frac{14}{15},1]$ and computes
$\sum_b(n_b/n)|\bar z_b-\bar c_b|$, skipping empty bins.  AURC sorts confidence from high to low; all equal-confidence items form an indivisible block, and we use the exact expected risk under a uniform random order within that block. HCE@$0.9=n^{-1}\sum_i\mathbf1[c_i\ge.9,z_i=0]$ is error mass in the entire evaluation set, not conditional error among accepted cases.

\paragraph{Calibration and uncertainty.}
Reader weights and standardization never see calibration, validation, test, or OOD images.  Scalar temperatures are fit by binary cross-entropy on the image-disjoint SLAKE-train calibration partition and then frozen.  No temperature is refit on any of the six OOD datasets. Reported uncertainty is the standard deviation over actor seeds $\{17,29,43\}$ as recorded by each frozen experiment artifact; the OOD6 aggregation uses the sample standard deviation. Paired claims additionally resample unique source images with replacement and retain all questions from a sampled image; grounding uses 10,000 percentile-bootstrap replicates.

\section{Baseline Implementation Details}
\label{app:baseline_protocols}

Two categories are distinguished throughout: a \emph{GRPO baseline} and an
\emph{Estimator baseline}. Base, RLVR, RLCR, DCPO, and VL-Calibration have their own generated answers.  P(True), self-probing, prompt ensemble, SAPLMA, BICR, sequence likelihood, self-consistency, semantic uncertainty, and DualRead all read the frozen RLVR actor's already committed natural answer.  Consequently the latter family shares RLVR accuracy exactly; only its confidence score can change.

\paragraph{Policy baselines.}
\textbf{Base} is the instruction-tuned checkpoint without task RL, evaluated under the same question and answer contract.  \textbf{RLVR} is the capability actor: the task term is $z$ and confidence receives no calibration reward. \textbf{RLCR} adds a pointwise proper-scoring term with $R=z-(c-z)^2$ (calibration weight one).  \textbf{DCPO} keeps its constitutive
segmented answer/confidence credit, hybrid group target $\gamma=0.5$,
asymmetric clipping $0.2/0.28$, and policy KL coefficient $10^{-3}$.
\textbf{VL-Calibration} keeps dual vision/reasoning confidences, fixed harmonic
aggregation, token shaping in $[0.9,1.1]$, the clean/augmented visual forward, and policy KL $10^{-4}$.  Its native five-part output requires 512 completion tokens, versus 256 for the shared short format; this budget follows the method's own output contract and is reported here as part of its cost. Actor exposure, batch size, group size, optimizer, grader,
training data, and three seeds otherwise follow Table~\ref{tab:hparams}.

\paragraph{Prompted self-check baselines.}
\textbf{P(True)} shows the frozen actor the image, question, and its own answer, asks for a one-word Yes/No judgment, and normalizes the two first-token probabilities.  It uses one follow-up forward pass and never changes the original answer. \textbf{Self-probing} instead requests one integer probability from 0 to 100; the last valid integer is used because the actor may reason before giving the score.  \textbf{Prompt ensemble} averages valid scores from
four fixed paraphrases of that request.  If none parses, the item is retained at $0.5$.  The first ensemble prompt is exactly the self-probing prompt, so the ensemble strictly contains rather than replaces the single-prompt baseline.

\paragraph{Hidden state baselines.}
\textbf{SAPLMA} is implemented as a mid-layer statement probe: one teacher-forced replay of the frozen completion, the last-token activation at $0.625$ of model depth (fixed before evaluation), and a $256$--$128$--$64$ ReLU classifier.  It uses the same image-disjoint
fit/calibration partitions and temperature protocol as DualRead.
\textbf{BICR} contrasts real-image and black-image states from the same actor. Its $1024$--$512$ MLP was selected once in a 50-trial seed-43 study; the architecture was then frozen and only weights were refit for each seed. \textbf{Sequence likelihood} is the geometric mean of generated answer-token probabilities, temperature-scaled only on SLAKE-train calibration images.

\paragraph{Sampling baselines.}
\textbf{Self-consistency (SC@$K$)} draws $K\in\{2,4,8\}$ completions at
temperature one and scores the deployed greedy answer by the fraction of
canonical sampled answers that match it; invalid samples count as
disagreement.  A positive-slope Platt map is fitted on SLAKE validation and
then frozen.  \textbf{Semantic uncertainty} partitions the same $K=8$
completions by the shared grader's canonical answer form, computes normalized
semantic entropy, and Platt-calibrates its negative on SLAKE validation.  No
external NLI model or target-domain labels enter the clustering.

\paragraph{Comparison costs.}
The RLVR token and sequence likelihood need no extra actor pass beyond natural generation; P(True) and self-probing need one follow-up pass; the prompt ensemble needs four; SAPLMA, BICR, and DualRead need one parallel
teacher-forced replay; SC and semantic uncertainty need $K$ stochastic
generations.

\subsection{Exact Prompt Templates}
\label{app:prompts}

\newcommand{\PromptBox}[1]{%
  \par\smallskip
  \noindent
  \begingroup
  \setlength{\fboxsep}{6pt}%
  \setlength{\fboxrule}{0.5pt}%
  \fbox{%
    \begin{minipage}{\dimexpr\linewidth-2\fboxsep-2\fboxrule\relax}
      \raggedright
      \textbf{Model-visible prompt}\par
      \vspace{2pt}%
      \hrule height 0.4pt
      \vspace{5pt}%
      \small\ttfamily
      \setlength{\parindent}{0pt}%
      #1%
    \end{minipage}%
  }%
  \endgroup
  \par\smallskip
}

We reproduce the exact model-visible instructions below.  \{QUESTION\},
\{ANSWER\}, and \{OPTION\_TEXT\} denote per-example substitutions, not literal
strings.  Qwen and MedGemma receive the same visible text.  Their processors
then apply the checkpoint's native chat template to one user message containing
the image followed by the text; we add no handwritten system message.  Vendor
chat delimiters are therefore backbone-specific and are not part of our
experimental manipulation.  Automatic wrapping inside each frame is
typographic only; explicit displayed line breaks are model-visible.

\paragraph{Shared natural-answer prompt.}
Base, RLVR, RLCR, DCPO, DualRead feature extraction, SAPLMA, BICR,
sequence-likelihood, SC@$K$, semantic uncertainty, and every SLAKE/VQA-RAD/
PathVQA natural generation use:
\PromptBox{%
Examine the medical image and answer the question.\\
Give concise image-grounded reasoning, then one final answer and one confidence.\\
Use exactly:\\
\textless reasoning\textgreater...\textless/reasoning\textgreater\\
\textless answer\textgreater...\textless/answer\textgreater\\
\textless confidence\textgreater{}NN\textless/confidence\textgreater\\
NN must be one integer from 0 to 100.\\[2pt]
Question: \{QUESTION\}
}
Training and inference instantiate this template through their respective data
pipelines, but after substituting the question they present byte-identical
model-visible user messages.  Greedy evaluation uses the instruction
unchanged.  SC and semantic uncertainty alter only the decoding rule
(temperature one and $K$ samples), not the prompt.

\paragraph{P(True).}
The image is shown again together with the frozen actor's already generated
answer.  One token is generated, and the reported score is the normalized
first-token probability of Yes against No:
\PromptBox{%
Question: \{QUESTION\}\\
Proposed answer: \{ANSWER\}\\
Is the proposed answer correct for this medical image? Respond with a single word:
Yes or No.
}
The original answer is never regenerated or modified by this follow-up.

\paragraph{Self-probing and prompt ensemble.}
The first template is the self-probing baseline.  Prompt ensemble evaluates all
four templates independently and averages the valid parsed probabilities:
\noindent\textit{Template 1 (self-probing).}
\PromptBox{%
Question: \{QUESTION\}\\
Proposed answer: \{ANSWER\}\\
How likely is the proposed answer to be correct for this medical image? Reply
with a single integer from 0 to 100.
}
\noindent\textit{Template 2.}
\PromptBox{%
Look at the image. A radiology question was asked: \{QUESTION\}\\
Someone answered: \{ANSWER\}\\
On a scale of 0 to 100, how confident are you that this answer is right? Reply
with the number only.
}
\noindent\textit{Template 3.}
\PromptBox{%
Medical image question: \{QUESTION\}\\
Candidate answer: \{ANSWER\}\\
Estimate the probability (0-100) that the candidate answer is correct. Answer
with just the number.
}
\noindent\textit{Template 4.}
\PromptBox{%
\{QUESTION\}\\
An assistant replied: \{ANSWER\}\\
Rate the chance that the reply is correct, from 0 (certainly wrong) to 100
(certainly right). Give only the integer.
}
Generation is greedy with at most 96 new tokens.  If a completion contains a
valid explicit \texttt{\textless confidence\textgreater NN\textless/confidence\textgreater}
block, that integer is used; otherwise the last integer in $[0,100]$ is used.
An invalid self-probe is retained at $0.5$; the ensemble averages its valid
members and returns $0.5$ only when none parses.

\paragraph{Option-visible multiple-choice prompt.}
For the PMC-VQA, OmniMedVQA, and ReXVQA, the shared
natural-answer instruction remains in force and the question field is:
\PromptBox{%
\{QUESTION\}\\
A. \{OPTION\_TEXT\_A\}\\
B. \{OPTION\_TEXT\_B\}\\
C. \{OPTION\_TEXT\_C\}\\
D. \{OPTION\_TEXT\_D\}\\[2pt]
Select the single best option. In \textless answer\textgreater, write the
complete text of the selected option exactly as shown; do not write only its
letter.
}
If a source question already contains all four labeled options, the option
block is not duplicated.  The appended selection instruction is identical.

\paragraph{VL-Calibration native prompt.}
VL-Calibration retains its method-specific five-part interface.  Its two
confidence values are nested inside one confidence block, which makes
the contract line 110 characters wide.  That line is a single unbroken line in the model-visible prompt; inside the box it is permitted to wrap at tag boundaries, and such a wrap inserts no character.
\PromptBox{%
Examine the medical image and answer the question.\\[2pt]
Reply with ALL FIVE parts below, in this order, using these exact tags:\\
\textless vision\textgreater{}what you actually see in the image, at most two
sentences\textless/vision\textgreater\\
\textless reasoning\textgreater{}why that supports your answer, at most two
sentences\textless/reasoning\textgreater\\
\textless answer\textgreater{}the final answer only: a single word or a short
phrase\textless/answer\textgreater\\
\textless analysis\textgreater{}one short sentence\textless/analysis\textgreater\\
\textless confidence\textgreater\allowbreak%
\textless vision\_confidence\textgreater\allowbreak%
N\textless/vision\_confidence\textgreater\allowbreak%
\textless reasoning\_confidence\textgreater\allowbreak%
N\textless/reasoning\_confidence\textgreater\allowbreak%
\textless/confidence\textgreater\\[2pt]
All five parts are required. Do not stop after the reasoning.\\
The last part must contain BOTH inner tags. Never write a single number such as\\
\textless confidence\textgreater\allowbreak%
9\textless/confidence\textgreater\ -- it must be\\
\textless confidence\textgreater\allowbreak%
\textless vision\_confidence\textgreater\allowbreak%
8\textless/vision\_confidence\textgreater\allowbreak%
\textless reasoning\_confidence\textgreater\allowbreak%
6\textless/reasoning\_confidence\textgreater\allowbreak%
\textless/confidence\textgreater,\\
where the first integer is how confident you are in what you saw and the second is how\\
confident you are in the reasoning. Both are integers from 0 to 10.\\
Write your own content, never copy this template.\\[2pt]
Question: \{QUESTION\}
}
Only this baseline uses the native five-part prompt.  Its two confidence values
are combined by the frozen harmonic aggregation described in
\S\ref{app:baseline_protocols}; all other policy actors use the shared prompt
above.  Fixed-prefix and strict-same-answer controls introduce no new
elicitation: they reuse the shared prompt and teacher-force the cached natural
prefix specified by each control.

\section{Properties and Scope of CCG-AUC}
\label{app:ccg_properties}
This section restates the construction of the main paper in the form used by the supplementary tables.  \emph{Donor} denotes the hard-negative image
$I^{\mathrm{hn}}_i$; \emph{source} denotes the original image $I_i$.
For each source item, candidate donors have the same normalized question,
a different canonical gold answer, a different image identifier, and a real
image from the same evaluation distribution.  The donor assignment is frozen before confidence scoring.  The actor is greedily decoded on the source image and on its donor image with the question unchanged.  With
$\varphi(d)=\mathbf 1[d>0]+\tfrac12\mathbf 1[d=0]$ and the actor-defined
decisive set
$\mathcal E=\{i:z_i^{\mathrm{real}}=1,z_i^{\mathrm{hn}}=0\}$, Natural CCG-AUC is
\begin{equation}
\operatorname{Natural\ CCG\mbox{-}AUC}
=
\frac{1}{|\mathcal E|}
\sum_{i\in\mathcal E}
\varphi\!\left(c_i^{\mathrm{real}}-c_i^{\mathrm{hn}}\right),
\label{eq:supp_ccg}
\end{equation}
which is Eq.~(10) of the main paper.  Strict CCG-AUC applies the same rule over
$\mathcal E_{\mathrm{strict}}=\{i\in\mathcal E: a_i^{\mathrm{real}}=a_i^{\mathrm{hn}}\neq\emptyset\}$,
i.e.\ it additionally requires equal, non-empty canonical generated answers
under the two images.  Both variants are reported throughout this supplement.
The primary DualRead/token comparison
uses the same RLVR actor and therefore exactly the same answers, labels, image
pairs, and decisive set.

\paragraph{Actor-side compatibility diagnostics.}
For completeness, we also report two answer-side quantities.  Let $\mathcal P$ be the complete frozen set of
source/hard-negative pairs, before conditioning on either confidence or actor
correctness.  We define
\begin{align}
\mathrm{VRS}
&=
\frac{1}{|\mathcal P|}
\sum_{i\in\mathcal P}
\left(z_i^{\mathrm{real}}-z_i^{\mathrm{hn}}\right)
=
\mathrm{Acc}^{\mathrm{real}}-\mathrm{Acc}^{\mathrm{hn}},
\\
\mathrm{VBR}
&=
\frac{1}{|\mathcal P|}
\sum_{i\in\mathcal P}
\mathbf 1[z_i^{\mathrm{real}}=1,z_i^{\mathrm{hn}}=0]
=
\frac{|\mathcal E|}{|\mathcal P|}.
\end{align}
VRS is the actor's net accuracy change and can be offset by pairs that flip in
the opposite direction; VBR is the incidence of the decisive failure event
used to define CCG eligibility.  Neither quantity contains a confidence score.
They are consequently identical for the RLVR token, PSR, P-Visual, DualRead,
and every other methods attached to the same frozen RLVR actor.  They are
reported as actor-side diagnostics and carry no information about any
confidence interface.

\paragraph{A parameter-, scale-, and model-free rule.}
CCG-AUC applies \emph{no post-processing or calibration} to the model's outputs.
It is \emph{parameter-free}: it fits nothing, uses no validation split, and has no
free coefficient or temperature.  It is \emph{scale-free}: $\varphi$ depends only on
the sign of the within-item difference $c^{\mathrm{real}}_i-c^{\mathrm{hn}}_i$, so the
score is invariant to any strictly increasing reparameterization of the confidence scale
(probability versus log-odds, arbitrary temperature).  It is \emph{model-agnostic}:
the rule consumes only per-item $(c,z)$ pairs and evaluates any confidence signal
---a verbalized token, hidden state, or sequence likelihood---as a
black box.  Because $c^{\mathrm{real}}$ and $c^{\mathrm{hn}}$ come from the same
estimator on the same item, comparisons across estimators need no cross-model
normalization.

\paragraph{Preconditions and transferability.}
These properties make the metric portable but place the burden on the
\emph{data construction}.  CCG-AUC requires that the dataset admit same-question
hard negatives: a repeated-question structure (the same normalized question asked
of multiple images), gold answers (to select a donor image with a different answer
and to compute $z$), and a shared grader.  It further requires statistical power:
we admit a cell only when $\mathcal{E}$ spans at least $30$ independent source and
donor images, and otherwise flag it underpowered and report an interval only.
Consequently CCG-AUC transfers to any VLM that exposes a confidence signal
and any VQA dataset with groupable QA structure and labels, without
retuning; conversely it is not constructible---or is underpowered---on datasets of
near-unique questions or without gold.  The construction uses \emph{real
in-distribution} hard negatives rather than blanked or shuffled images, so the
substituted input stays inside the evaluation distribution and the measured
response is not a response to input novelty.

\paragraph{Inference and scope.}
Because multiple questions can share a source image, we use a paired
source-image-cluster bootstrap for confidence intervals and paired
system comparisons.  CCG-AUC measures whether the confidence score tracks
answer-decisive visual evidence; it is not pixel localization, clinical causal
attribution, or a claim that the actor's \emph{answer} is more grounded (the latter
is a separate actor-level precondition).

\section{Paired Bootstrap Contrasts}
\label{app:cascade_ci}

Table~\ref{tab:paired} summarizes paired image-cluster bootstrap contrasts for
P-Visual versus P-State on SLAKE validation and for DualRead versus RLCR on
the ID and OOD evaluation splits.
\begin{table*}[!t]
\centering
\small
\setlength{\tabcolsep}{4pt}
\renewcommand{\arraystretch}{1.05}
\begin{tabular*}{\textwidth}{@{\extracolsep{\fill}}llrr@{}}
\toprule
Backbone & Contrast (split) & $\Delta$AUROC & $\Delta$Brier\\
\midrule
\multirow{3}{*}{Qwen3VL-2B}
& P-Visual $-$ P-State (val.) & $+.0183_{\pm.0076}$ & $-.0157_{\pm.0059}$\\
& DualRead $-$ RLCR (ID) & $+.130_{\pm.015}$ & $+.067_{\pm.013}$\\
& DualRead $-$ RLCR (OOD) & $+.020_{\pm.012}$ & $+.077_{\pm.011}$\\
\midrule
\multirow{3}{*}{MedGemma1.5-4B}
& P-Visual $-$ P-State (val.) & $+.0114_{\pm.0053}$ & ---\\
& DualRead $-$ RLCR (ID) & $+.205_{\pm.024}$ & $+.110_{\pm.020}$\\
& DualRead $-$ RLCR (OOD) & $+.075_{\pm.023}$ & $+.102_{\pm.010}$\\
\bottomrule
\end{tabular*}
\caption{Paired image-cluster bootstrap contrasts (mean $\pm$ std).}
\label{tab:paired}
\end{table*}

\section{Extended Reliability Results}
\label{app:extended}

\subsection{Calibration metrics corresponding to the main table}
\label{app:main_calibration}

The main-paper table prioritizes discrimination and grounding and therefore
omits two calibration columns. Tables~\ref{tab:ood_acc} and~\ref{tab:calib}
restore ECE-15 and Brier alongside Acc./AUROC. The SLAKE columns are
computed on the ID test set. For six OOD test sets, each metric is first computed separately
on VQA-RAD, PathVQA, PMC-VQA, OmniMedVQA, ReXVQA, and the closed subset of
VQA-Med 2019; the six values are then averaged with equal domain weight for
each seed, followed by the across-seed summary. Base is one frozen zero-shot run;
other rows summarize actor seeds $\{17,29,43\}$.

\begin{table*}[!t]
\centering
\footnotesize
\setlength{\tabcolsep}{2pt}
\renewcommand{\arraystretch}{.96}
\begin{tabular}{@{}llccc@{}}
\toprule
Backbone & Method & SLAKE & VQA-RAD & PathVQA\\
& & \multicolumn{3}{c}{Acc./AUROC (top); ECE-15/Brier (bottom)}\\
\midrule
\multirow{11}{*}{\rotatebox[origin=c]{90}{Qwen3VL-2B}}
& Base & \shortstack{\ensuremath{.443}\,/\,\ensuremath{.579}\\\ensuremath{.539}\,/\,\ensuremath{.515}} & \shortstack{\ensuremath{.340}\,/\,\ensuremath{.644}\\\ensuremath{.550}\,/\,\ensuremath{.510}} & \shortstack{\ensuremath{.295}\,/\,\ensuremath{.731}\\\ensuremath{.517}\,/\,\ensuremath{.433}}\\
& RLCR & \shortstack{\ensuremath{.608_{\pm\!.009}}\,/\,\ensuremath{.822_{\pm\!.014}}\\\ensuremath{.150_{\pm\!.012}}\,/\,\ensuremath{.150_{\pm\!.012}}} & \shortstack{\ensuremath{.440_{\pm\!.009}}\,/\,\ensuremath{.709_{\pm\!.006}}\\\ensuremath{.312_{\pm\!.010}}\,/\,\ensuremath{.312_{\pm\!.010}}} & \shortstack{\ensuremath{.330_{\pm\!.007}}\,/\,\ensuremath{.820_{\pm\!.008}}\\\ensuremath{.218_{\pm\!.019}}\,/\,\ensuremath{.218_{\pm\!.019}}}\\
& DCPO & \shortstack{\ensuremath{.640_{\pm\!.003}}\,/\,\ensuremath{.789_{\pm\!.007}}\\\ensuremath{.190_{\pm\!.016}}\,/\,\ensuremath{.189_{\pm\!.016}}} & \shortstack{\ensuremath{.448_{\pm\!.011}}\,/\,\ensuremath{.731_{\pm\!.010}}\\\ensuremath{.277_{\pm\!.003}}\,/\,\ensuremath{.277_{\pm\!.003}}} & \shortstack{\ensuremath{.335_{\pm\!.010}}\,/\,\ensuremath{.800_{\pm\!.019}}\\\ensuremath{.214_{\pm\!.009}}\,/\,\ensuremath{.214_{\pm\!.009}}}\\
& VL-Cal. & \shortstack{\ensuremath{.591_{\pm\!.010}}\,/\,\ensuremath{.756_{\pm\!.026}}\\\ensuremath{.183_{\pm\!.048}}\,/\,\ensuremath{.207_{\pm\!.036}}} & \shortstack{\ensuremath{.443_{\pm\!.017}}\,/\,\ensuremath{.709_{\pm\!.006}}\\\ensuremath{.282_{\pm\!.057}}\,/\,\ensuremath{.294_{\pm\!.034}}} & \shortstack{\ensuremath{.332_{\pm\!.012}}\,/\,\ensuremath{.775_{\pm\!.007}}\\\ensuremath{.276_{\pm\!.085}}\,/\,\ensuremath{.273_{\pm\!.060}}}\\
& GRPO (RLVR) & \shortstack{\ensuremath{.639_{\pm\!.009}}\,/\,\ensuremath{.535_{\pm\!.059}}\\\ensuremath{.359_{\pm\!.012}}\,/\,\ensuremath{.365_{\pm\!.005}}} & \shortstack{\ensuremath{.453_{\pm\!.006}}\,/\,\ensuremath{.490_{\pm\!.019}}\\\ensuremath{.538_{\pm\!.017}}\,/\,\ensuremath{.531_{\pm\!.028}}} & \shortstack{\ensuremath{.339_{\pm\!.003}}\,/\,\ensuremath{.465_{\pm\!.061}}\\\ensuremath{.634_{\pm\!.044}}\,/\,\ensuremath{.621_{\pm\!.067}}}\\
& P(True) & \shortstack{\ensuremath{.639_{\pm\!.009}}\,/\,\ensuremath{.609_{\pm\!.012}}\\\ensuremath{.381_{\pm\!.013}}\,/\,\ensuremath{.385_{\pm\!.011}}} & \shortstack{\ensuremath{.453_{\pm\!.006}}\,/\,\ensuremath{.444_{\pm\!.005}}\\\ensuremath{.433_{\pm\!.016}}\,/\,\ensuremath{.460_{\pm\!.012}}} & \shortstack{\ensuremath{.339_{\pm\!.003}}\,/\,\ensuremath{.344_{\pm\!.019}}\\\ensuremath{.504_{\pm\!.028}}\,/\,\ensuremath{.530_{\pm\!.025}}}\\
& Self-prob. & \shortstack{\ensuremath{.639_{\pm\!.009}}\,/\,\ensuremath{.517_{\pm\!.033}}\\\ensuremath{.408_{\pm\!.023}}\,/\,\ensuremath{.411_{\pm\!.018}}} & \shortstack{\ensuremath{.453_{\pm\!.006}}\,/\,\ensuremath{.447_{\pm\!.020}}\\\ensuremath{.567_{\pm\!.030}}\,/\,\ensuremath{.561_{\pm\!.034}}} & \shortstack{\ensuremath{.339_{\pm\!.003}}\,/\,\ensuremath{.379_{\pm\!.009}}\\\ensuremath{.653_{\pm\!.032}}\,/\,\ensuremath{.646_{\pm\!.040}}}\\
& Prompt ens. & \shortstack{\ensuremath{.639_{\pm\!.009}}\,/\,\ensuremath{.534_{\pm\!.034}}\\\ensuremath{.348_{\pm\!.003}}\,/\,\ensuremath{.362_{\pm\!.002}}} & \shortstack{\ensuremath{.453_{\pm\!.006}}\,/\,\ensuremath{.459_{\pm\!.012}}\\\ensuremath{.457_{\pm\!.009}}\,/\,\ensuremath{.483_{\pm\!.013}}} & \shortstack{\ensuremath{.339_{\pm\!.003}}\,/\,\ensuremath{.372_{\pm\!.005}}\\\ensuremath{.599_{\pm\!.027}}\,/\,\ensuremath{.600_{\pm\!.042}}}\\
& SAPLMA & \shortstack{\ensuremath{.639_{\pm\!.009}}\,/\,\ensuremath{.942_{\pm\!.004}}\\\ensuremath{.042_{\pm\!.007}}\,/\,\ensuremath{.095_{\pm\!.005}}} & \shortstack{\ensuremath{.453_{\pm\!.006}}\,/\,\ensuremath{\mathbf{.747}_{\pm\!.008}}\\\ensuremath{\mathbf{.118}_{\pm\!.012}}\,/\,\ensuremath{\mathbf{.215}_{\pm\!.008}}} & \shortstack{\ensuremath{.339_{\pm\!.003}}\,/\,\ensuremath{\mathbf{.833}_{\pm\!.024}}\\\ensuremath{.157_{\pm\!.060}}\,/\,\ensuremath{\mathbf{.189}_{\pm\!.038}}}\\
& BICR & \shortstack{\ensuremath{.639_{\pm\!.009}}\,/\,\ensuremath{.934_{\pm\!.002}}\\\ensuremath{.035_{\pm\!.015}}\,/\,\ensuremath{.099_{\pm\!.003}}} & \shortstack{\ensuremath{.453_{\pm\!.006}}\,/\,\ensuremath{.686_{\pm\!.019}}\\\ensuremath{.190_{\pm\!.050}}\,/\,\ensuremath{.266_{\pm\!.020}}} & \shortstack{\ensuremath{.339_{\pm\!.003}}\,/\,\ensuremath{.704_{\pm\!.024}}\\\ensuremath{\mathbf{.143}_{\pm\!.052}}\,/\,\ensuremath{.224_{\pm\!.024}}}\\
& DualRead & \shortstack{\ensuremath{.639_{\pm\!.009}}\,/\,\ensuremath{\mathbf{.952}_{\pm\!.004}}\\\ensuremath{\mathbf{.031}_{\pm\!.007}}\,/\,\ensuremath{\mathbf{.083}_{\pm\!.002}}} & \shortstack{\ensuremath{.453_{\pm\!.006}}\,/\,\ensuremath{.731_{\pm\!.008}}\\\ensuremath{.160_{\pm\!.017}}\,/\,\ensuremath{.233_{\pm\!.008}}} & \shortstack{\ensuremath{.339_{\pm\!.003}}\,/\,\ensuremath{.784_{\pm\!.078}}\\\ensuremath{.170_{\pm\!.094}}\,/\,\ensuremath{.224_{\pm\!.073}}}\\
\midrule
\multirow{11}{*}{\rotatebox[origin=c]{90}{MedGemma1.5-4B}}
& Base & \shortstack{\ensuremath{.478}\,/\,\ensuremath{.615}\\\ensuremath{.430}\,/\,\ensuremath{.429}} & \shortstack{\ensuremath{.463}\,/\,\ensuremath{.552}\\\ensuremath{.373}\,/\,\ensuremath{.402}} & \shortstack{\ensuremath{.294}\,/\,\ensuremath{.503}\\\ensuremath{.585}\,/\,\ensuremath{.566}}\\
& RLCR & \shortstack{\ensuremath{.657_{\pm\!.027}}\,/\,\ensuremath{.771_{\pm\!.021}}\\\ensuremath{.172_{\pm\!.019}}\,/\,\ensuremath{.172_{\pm\!.019}}} & \shortstack{\ensuremath{.468_{\pm\!.039}}\,/\,\ensuremath{.676_{\pm\!.044}}\\\ensuremath{.340_{\pm\!.026}}\,/\,\ensuremath{.340_{\pm\!.026}}} & \shortstack{\ensuremath{.351_{\pm\!.011}}\,/\,\ensuremath{.663_{\pm\!.067}}\\\ensuremath{.429_{\pm\!.088}}\,/\,\ensuremath{.429_{\pm\!.088}}}\\
& DCPO & \shortstack{\ensuremath{.678_{\pm\!.003}}\,/\,\ensuremath{.752_{\pm\!.018}}\\\ensuremath{.187_{\pm\!.011}}\,/\,\ensuremath{.187_{\pm\!.012}}} & \shortstack{\ensuremath{.445_{\pm\!.078}}\,/\,\ensuremath{.691_{\pm\!.038}}\\\ensuremath{.342_{\pm\!.080}}\,/\,\ensuremath{.342_{\pm\!.081}}} & \shortstack{\ensuremath{.318_{\pm\!.060}}\,/\,\ensuremath{.765_{\pm\!.027}}\\\ensuremath{.313_{\pm\!.027}}\,/\,\ensuremath{.313_{\pm\!.026}}}\\
& VL-Cal. & \shortstack{\ensuremath{.587_{\pm\!.005}}\,/\,\ensuremath{.851_{\pm\!.005}}\\\ensuremath{.085_{\pm\!.028}}\,/\,\ensuremath{.127_{\pm\!.015}}} & \shortstack{\ensuremath{.477_{\pm\!.013}}\,/\,\ensuremath{\mathbf{.754}_{\pm\!.012}}\\\ensuremath{.161_{\pm\!.068}}\,/\,\ensuremath{\mathbf{.215}_{\pm\!.027}}} & \shortstack{\ensuremath{.343_{\pm\!.015}}\,/\,\ensuremath{.713_{\pm\!.031}}\\\ensuremath{.202_{\pm\!.040}}\,/\,\ensuremath{.234_{\pm\!.013}}}\\
& GRPO (RLVR) & \shortstack{\ensuremath{.677_{\pm\!.011}}\,/\,\ensuremath{.506_{\pm\!.011}}\\\ensuremath{.229_{\pm\!.100}}\,/\,\ensuremath{.278_{\pm\!.051}}} & \shortstack{\ensuremath{.498_{\pm\!.011}}\,/\,\ensuremath{.504_{\pm\!.007}}\\\ensuremath{.316_{\pm\!.243}}\,/\,\ensuremath{.389_{\pm\!.124}}} & \shortstack{\ensuremath{.354_{\pm\!.010}}\,/\,\ensuremath{.503_{\pm\!.004}}\\\ensuremath{.460_{\pm\!.236}}\,/\,\ensuremath{.477_{\pm\!.195}}}\\
& P(True) & \shortstack{\ensuremath{.677_{\pm\!.011}}\,/\,\ensuremath{.574_{\pm\!.026}}\\\ensuremath{.362_{\pm\!.019}}\,/\,\ensuremath{.362_{\pm\!.019}}} & \shortstack{\ensuremath{.498_{\pm\!.011}}\,/\,\ensuremath{.558_{\pm\!.013}}\\\ensuremath{.478_{\pm\!.023}}\,/\,\ensuremath{.478_{\pm\!.024}}} & \shortstack{\ensuremath{.354_{\pm\!.010}}\,/\,\ensuremath{.439_{\pm\!.045}}\\\ensuremath{.655_{\pm\!.013}}\,/\,\ensuremath{.654_{\pm\!.011}}}\\
& Self-prob. & \shortstack{\ensuremath{.677_{\pm\!.011}}\,/\,\ensuremath{.597_{\pm\!.110}}\\\ensuremath{.289_{\pm\!.039}}\,/\,\ensuremath{.298_{\pm\!.037}}} & \shortstack{\ensuremath{.498_{\pm\!.011}}\,/\,\ensuremath{.551_{\pm\!.055}}\\\ensuremath{.437_{\pm\!.036}}\,/\,\ensuremath{.445_{\pm\!.039}}} & \shortstack{\ensuremath{.354_{\pm\!.010}}\,/\,\ensuremath{.444_{\pm\!.025}}\\\ensuremath{.580_{\pm\!.020}}\,/\,\ensuremath{.592_{\pm\!.029}}}\\
& Prompt ens. & \shortstack{\ensuremath{.677_{\pm\!.011}}\,/\,\ensuremath{.626_{\pm\!.073}}\\\ensuremath{.242_{\pm\!.047}}\,/\,\ensuremath{.272_{\pm\!.034}}} & \shortstack{\ensuremath{.498_{\pm\!.011}}\,/\,\ensuremath{.553_{\pm\!.013}}\\\ensuremath{.397_{\pm\!.066}}\,/\,\ensuremath{.415_{\pm\!.054}}} & \shortstack{\ensuremath{.354_{\pm\!.010}}\,/\,\ensuremath{.454_{\pm\!.006}}\\\ensuremath{.560_{\pm\!.049}}\,/\,\ensuremath{.554_{\pm\!.061}}}\\
& SAPLMA & \shortstack{\ensuremath{.677_{\pm\!.011}}\,/\,\ensuremath{.942_{\pm\!.011}}\\\ensuremath{\mathbf{.035}_{\pm\!.004}}\,/\,\ensuremath{.087_{\pm\!.010}}} & \shortstack{\ensuremath{.498_{\pm\!.011}}\,/\,\ensuremath{.736_{\pm\!.010}}\\\ensuremath{\mathbf{.138}_{\pm\!.020}}\,/\,\ensuremath{.228_{\pm\!.005}}} & \shortstack{\ensuremath{.354_{\pm\!.010}}\,/\,\ensuremath{.743_{\pm\!.059}}\\\ensuremath{.162_{\pm\!.077}}\,/\,\ensuremath{.238_{\pm\!.055}}}\\
& BICR & \shortstack{\ensuremath{.677_{\pm\!.011}}\,/\,\ensuremath{.930_{\pm\!.007}}\\\ensuremath{.040_{\pm\!.007}}\,/\,\ensuremath{.100_{\pm\!.007}}} & \shortstack{\ensuremath{.498_{\pm\!.011}}\,/\,\ensuremath{.723_{\pm\!.002}}\\\ensuremath{.271_{\pm\!.039}}\,/\,\ensuremath{.288_{\pm\!.023}}} & \shortstack{\ensuremath{.354_{\pm\!.010}}\,/\,\ensuremath{.723_{\pm\!.091}}\\\ensuremath{.176_{\pm\!.071}}\,/\,\ensuremath{.239_{\pm\!.055}}}\\
& DualRead & \shortstack{\ensuremath{.677_{\pm\!.011}}\,/\,\ensuremath{\mathbf{.947}_{\pm\!.005}}\\\ensuremath{.038_{\pm\!.008}}\,/\,\ensuremath{\mathbf{.082}_{\pm\!.002}}} & \shortstack{\ensuremath{.498_{\pm\!.011}}\,/\,\ensuremath{\mathbf{.754}_{\pm\!.016}}\\\ensuremath{.142_{\pm\!.030}}\,/\,\ensuremath{.219_{\pm\!.017}}} & \shortstack{\ensuremath{.354_{\pm\!.010}}\,/\,\ensuremath{\mathbf{.807}_{\pm\!.034}}\\\ensuremath{\mathbf{.144}_{\pm\!.046}}\,/\,\ensuremath{\mathbf{.196}_{\pm\!.025}}}\\
\bottomrule
\end{tabular}
\caption{Reliability on SLAKE, VQA-RAD, and PathVQA. Cells show Acc./AUROC over ECE-15/Brier; bold excludes Acc.}
\label{tab:ood_acc}
\end{table*}

\begin{table*}[!t]
\centering
\footnotesize
\setlength{\tabcolsep}{2pt}
\renewcommand{\arraystretch}{.96}
\setlength{\tabcolsep}{1.5pt}
\begin{tabular}{@{}llccccc@{}}
\toprule
Backbone & Method & PMC-VQA & OmniMed & ReXVQA & VQA-Med19-C & OOD6\\
& & \multicolumn{5}{c}{Acc./AUROC (top); ECE-15/Brier (bottom)}\\
\midrule
\multirow{11}{*}{\rotatebox[origin=c]{90}{Qwen3VL-2B}}
& Base & \shortstack{\ensuremath{.449}\,/\,\ensuremath{.681}\\\ensuremath{.468}\,/\,\ensuremath{.437}} & \shortstack{\ensuremath{.714}\,/\,\ensuremath{.705}\\\ensuremath{.225}\,/\,\ensuremath{.228}} & \shortstack{\ensuremath{.493}\,/\,\ensuremath{\mathbf{.617}}\\\ensuremath{.441}\,/\,\ensuremath{.426}} & \shortstack{\ensuremath{.406}\,/\,\ensuremath{\mathbf{.799}}\\\ensuremath{.429}\,/\,\ensuremath{.361}} & \shortstack{\ensuremath{.450}\,/\,\ensuremath{.696}\\\ensuremath{.438}\,/\,\ensuremath{.399}}\\
& RLCR & \shortstack{\ensuremath{.511_{\pm\!.010}}\,/\,\ensuremath{.563_{\pm\!.005}}\\\ensuremath{.433_{\pm\!.001}}\,/\,\ensuremath{.432_{\pm\!.001}}} & \shortstack{\ensuremath{.709_{\pm\!.003}}\,/\,\ensuremath{.561_{\pm\!.020}}\\\ensuremath{.289_{\pm\!.005}}\,/\,\ensuremath{.289_{\pm\!.005}}} & \shortstack{\ensuremath{.285_{\pm\!.044}}\,/\,\ensuremath{.500_{\pm\!.014}}\\\ensuremath{.615_{\pm\!.022}}\,/\,\ensuremath{.615_{\pm\!.022}}} & \shortstack{\ensuremath{.677_{\pm\!.039}}\,/\,\ensuremath{.523_{\pm\!.024}}\\\ensuremath{.312_{\pm\!.031}}\,/\,\ensuremath{.312_{\pm\!.031}}} & \shortstack{\ensuremath{.492_{\pm\!.012}}\,/\,\ensuremath{.613_{\pm\!.008}}\\\ensuremath{.363_{\pm\!.003}}\,/\,\ensuremath{.363_{\pm\!.003}}}\\
& DCPO & \shortstack{\ensuremath{.501_{\pm\!.005}}\,/\,\ensuremath{.588_{\pm\!.023}}\\\ensuremath{.413_{\pm\!.024}}\,/\,\ensuremath{.411_{\pm\!.025}}} & \shortstack{\ensuremath{.700_{\pm\!.012}}\,/\,\ensuremath{.613_{\pm\!.011}}\\\ensuremath{.289_{\pm\!.021}}\,/\,\ensuremath{.288_{\pm\!.021}}} & \shortstack{\ensuremath{.275_{\pm\!.030}}\,/\,\ensuremath{.488_{\pm\!.022}}\\\ensuremath{.572_{\pm\!.034}}\,/\,\ensuremath{.571_{\pm\!.034}}} & \shortstack{\ensuremath{.688_{\pm\!.031}}\,/\,\ensuremath{.512_{\pm\!.001}}\\\ensuremath{.323_{\pm\!.018}}\,/\,\ensuremath{.323_{\pm\!.018}}} & \shortstack{\ensuremath{.491_{\pm\!.004}}\,/\,\ensuremath{.622_{\pm\!.005}}\\\ensuremath{.348_{\pm\!.006}}\,/\,\ensuremath{.347_{\pm\!.007}}}\\
& VL-Cal. & \shortstack{\ensuremath{.480_{\pm\!.008}}\,/\,\ensuremath{.622_{\pm\!.029}}\\\ensuremath{.343_{\pm\!.027}}\,/\,\ensuremath{.354_{\pm\!.025}}} & \shortstack{\ensuremath{.716_{\pm\!.010}}\,/\,\ensuremath{.648_{\pm\!.047}}\\\ensuremath{.240_{\pm\!.084}}\,/\,\ensuremath{.276_{\pm\!.070}}} & \shortstack{\ensuremath{.352_{\pm\!.025}}\,/\,\ensuremath{.508_{\pm\!.008}}\\\ensuremath{.444_{\pm\!.067}}\,/\,\ensuremath{.436_{\pm\!.066}}} & \shortstack{\ensuremath{.661_{\pm\!.039}}\,/\,\ensuremath{.538_{\pm\!.053}}\\\ensuremath{.253_{\pm\!.028}}\,/\,\ensuremath{.278_{\pm\!.017}}} & \shortstack{\ensuremath{.497_{\pm\!.010}}\,/\,\ensuremath{.633_{\pm\!.020}}\\\ensuremath{.306_{\pm\!.024}}\,/\,\ensuremath{.318_{\pm\!.016}}}\\
& GRPO (RLVR) & \shortstack{\ensuremath{.514_{\pm\!.006}}\,/\,\ensuremath{.532_{\pm\!.045}}\\\ensuremath{.409_{\pm\!.132}}\,/\,\ensuremath{.427_{\pm\!.099}}} & \shortstack{\ensuremath{.709_{\pm\!.008}}\,/\,\ensuremath{.567_{\pm\!.105}}\\\ensuremath{.244_{\pm\!.080}}\,/\,\ensuremath{.269_{\pm\!.037}}} & \shortstack{\ensuremath{.283_{\pm\!.007}}\,/\,\ensuremath{.492_{\pm\!.012}}\\\ensuremath{.642_{\pm\!.134}}\,/\,\ensuremath{.629_{\pm\!.156}}} & \shortstack{\ensuremath{.667_{\pm\!.018}}\,/\,\ensuremath{.495_{\pm\!.008}}\\\ensuremath{.368_{\pm\!.070}}\,/\,\ensuremath{.373_{\pm\!.079}}} & \shortstack{\ensuremath{.494_{\pm\!.004}}\,/\,\ensuremath{.507_{\pm\!.008}}\\\ensuremath{.472_{\pm\!.056}}\,/\,\ensuremath{.475_{\pm\!.051}}}\\
& P(True) & \shortstack{\ensuremath{.514_{\pm\!.006}}\,/\,\ensuremath{\mathbf{.689}_{\pm\!.010}}\\\ensuremath{.388_{\pm\!.021}}\,/\,\ensuremath{.396_{\pm\!.014}}} & \shortstack{\ensuremath{.709_{\pm\!.008}}\,/\,\ensuremath{\mathbf{.787}_{\pm\!.008}}\\\ensuremath{.222_{\pm\!.002}}\,/\,\ensuremath{.232_{\pm\!.003}}} & \shortstack{\ensuremath{.283_{\pm\!.007}}\,/\,\ensuremath{.591_{\pm\!.055}}\\\ensuremath{.522_{\pm\!.045}}\,/\,\ensuremath{.515_{\pm\!.049}}} & \shortstack{\ensuremath{.667_{\pm\!.018}}\,/\,\ensuremath{.471_{\pm\!.032}}\\\ensuremath{.440_{\pm\!.033}}\,/\,\ensuremath{.421_{\pm\!.037}}} & \shortstack{\ensuremath{.494_{\pm\!.004}}\,/\,\ensuremath{.555_{\pm\!.010}}\\\ensuremath{.418_{\pm\!.018}}\,/\,\ensuremath{.426_{\pm\!.014}}}\\
& Self-prob. & \shortstack{\ensuremath{.514_{\pm\!.006}}\,/\,\ensuremath{.600_{\pm\!.034}}\\\ensuremath{.380_{\pm\!.055}}\,/\,\ensuremath{.387_{\pm\!.045}}} & \shortstack{\ensuremath{.709_{\pm\!.008}}\,/\,\ensuremath{.666_{\pm\!.031}}\\\ensuremath{.205_{\pm\!.037}}\,/\,\ensuremath{.231_{\pm\!.014}}} & \shortstack{\ensuremath{.283_{\pm\!.007}}\,/\,\ensuremath{.441_{\pm\!.066}}\\\ensuremath{.555_{\pm\!.091}}\,/\,\ensuremath{.559_{\pm\!.103}}} & \shortstack{\ensuremath{.667_{\pm\!.018}}\,/\,\ensuremath{.513_{\pm\!.029}}\\\ensuremath{.511_{\pm\!.010}}\,/\,\ensuremath{.508_{\pm\!.007}}} & \shortstack{\ensuremath{.494_{\pm\!.004}}\,/\,\ensuremath{.508_{\pm\!.028}}\\\ensuremath{.479_{\pm\!.039}}\,/\,\ensuremath{.482_{\pm\!.038}}}\\
& Prompt ens. & \shortstack{\ensuremath{.514_{\pm\!.006}}\,/\,\ensuremath{.625_{\pm\!.015}}\\\ensuremath{.362_{\pm\!.052}}\,/\,\ensuremath{.374_{\pm\!.038}}} & \shortstack{\ensuremath{.709_{\pm\!.008}}\,/\,\ensuremath{.717_{\pm\!.019}}\\\ensuremath{.191_{\pm\!.034}}\,/\,\ensuremath{.222_{\pm\!.012}}} & \shortstack{\ensuremath{.283_{\pm\!.007}}\,/\,\ensuremath{.402_{\pm\!.071}}\\\ensuremath{.556_{\pm\!.069}}\,/\,\ensuremath{.552_{\pm\!.077}}} & \shortstack{\ensuremath{.667_{\pm\!.018}}\,/\,\ensuremath{.538_{\pm\!.032}}\\\ensuremath{.276_{\pm\!.040}}\,/\,\ensuremath{.297_{\pm\!.042}}} & \shortstack{\ensuremath{.494_{\pm\!.004}}\,/\,\ensuremath{.519_{\pm\!.019}}\\\ensuremath{.407_{\pm\!.027}}\,/\,\ensuremath{.421_{\pm\!.023}}}\\
& SAPLMA & \shortstack{\ensuremath{.514_{\pm\!.006}}\,/\,\ensuremath{.645_{\pm\!.005}}\\\ensuremath{.289_{\pm\!.032}}\,/\,\ensuremath{.328_{\pm\!.022}}} & \shortstack{\ensuremath{.709_{\pm\!.008}}\,/\,\ensuremath{.755_{\pm\!.006}}\\\ensuremath{\mathbf{.162}_{\pm\!.030}}\,/\,\ensuremath{\mathbf{.208}_{\pm\!.011}}} & \shortstack{\ensuremath{.283_{\pm\!.007}}\,/\,\ensuremath{.537_{\pm\!.033}}\\\ensuremath{.317_{\pm\!.115}}\,/\,\ensuremath{.344_{\pm\!.090}}} & \shortstack{\ensuremath{.667_{\pm\!.018}}\,/\,\ensuremath{.666_{\pm\!.025}}\\\ensuremath{\mathbf{.210}_{\pm\!.017}}\,/\,\ensuremath{\mathbf{.231}_{\pm\!.015}}} & \shortstack{\ensuremath{.494_{\pm\!.004}}\,/\,\ensuremath{\mathbf{.697}_{\pm\!.006}}\\\ensuremath{\mathbf{.209}_{\pm\!.042}}\,/\,\ensuremath{\mathbf{.253}_{\pm\!.027}}}\\
& BICR & \shortstack{\ensuremath{.514_{\pm\!.006}}\,/\,\ensuremath{.619_{\pm\!.015}}\\\ensuremath{.443_{\pm\!.012}}\,/\,\ensuremath{.445_{\pm\!.011}}} & \shortstack{\ensuremath{.709_{\pm\!.008}}\,/\,\ensuremath{.643_{\pm\!.008}}\\\ensuremath{.282_{\pm\!.014}}\,/\,\ensuremath{.285_{\pm\!.012}}} & \shortstack{\ensuremath{.283_{\pm\!.007}}\,/\,\ensuremath{.513_{\pm\!.014}}\\\ensuremath{.689_{\pm\!.016}}\,/\,\ensuremath{.679_{\pm\!.021}}} & \shortstack{\ensuremath{.667_{\pm\!.018}}\,/\,\ensuremath{.700_{\pm\!.048}}\\\ensuremath{.241_{\pm\!.017}}\,/\,\ensuremath{.255_{\pm\!.013}}} & \shortstack{\ensuremath{.494_{\pm\!.004}}\,/\,\ensuremath{.644_{\pm\!.005}}\\\ensuremath{.332_{\pm\!.011}}\,/\,\ensuremath{.359_{\pm\!.007}}}\\
& DualRead & \shortstack{\ensuremath{.514_{\pm\!.006}}\,/\,\ensuremath{.612_{\pm\!.011}}\\\ensuremath{\mathbf{.249}_{\pm\!.024}}\,/\,\ensuremath{\mathbf{.311}_{\pm\!.010}}} & \shortstack{\ensuremath{.709_{\pm\!.008}}\,/\,\ensuremath{.667_{\pm\!.014}}\\\ensuremath{.166_{\pm\!.027}}\,/\,\ensuremath{.229_{\pm\!.014}}} & \shortstack{\ensuremath{.283_{\pm\!.007}}\,/\,\ensuremath{.520_{\pm\!.040}}\\\ensuremath{\mathbf{.305}_{\pm\!.020}}\,/\,\ensuremath{\mathbf{.342}_{\pm\!.020}}} & \shortstack{\ensuremath{.667_{\pm\!.018}}\,/\,\ensuremath{.683_{\pm\!.034}}\\\ensuremath{.253_{\pm\!.027}}\,/\,\ensuremath{.262_{\pm\!.023}}} & \shortstack{\ensuremath{.494_{\pm\!.004}}\,/\,\ensuremath{.666_{\pm\!.021}}\\\ensuremath{.217_{\pm\!.028}}\,/\,\ensuremath{.267_{\pm\!.021}}}\\
\midrule
\multirow{11}{*}{\rotatebox[origin=c]{90}{MedGemma1.5-4B}}
& Base & \shortstack{\ensuremath{.365}\,/\,\ensuremath{\mathbf{.698}}\\\ensuremath{.476}\,/\,\ensuremath{.418}} & \shortstack{\ensuremath{.586}\,/\,\ensuremath{.641}\\\ensuremath{.247}\,/\,\ensuremath{.276}} & \shortstack{\ensuremath{.693}\,/\,\ensuremath{.564}\\\ensuremath{.188}\,/\,\ensuremath{.242}} & \shortstack{\ensuremath{.562}\,/\,\ensuremath{.581}\\\ensuremath{.357}\,/\,\ensuremath{.365}} & \shortstack{\ensuremath{.494}\,/\,\ensuremath{.590}\\\ensuremath{.371}\,/\,\ensuremath{.378}}\\
& RLCR & \shortstack{\ensuremath{.490_{\pm\!.010}}\,/\,\ensuremath{.590_{\pm\!.005}}\\\ensuremath{.415_{\pm\!.002}}\,/\,\ensuremath{.414_{\pm\!.002}}} & \shortstack{\ensuremath{.653_{\pm\!.043}}\,/\,\ensuremath{.609_{\pm\!.042}}\\\ensuremath{.377_{\pm\!.034}}\,/\,\ensuremath{.377_{\pm\!.034}}} & \shortstack{\ensuremath{.858_{\pm\!.010}}\,/\,\ensuremath{.578_{\pm\!.010}}\\\ensuremath{.308_{\pm\!.024}}\,/\,\ensuremath{.308_{\pm\!.024}}} & \shortstack{\ensuremath{.635_{\pm\!.033}}\,/\,\ensuremath{.492_{\pm\!.013}}\\\ensuremath{.375_{\pm\!.016}}\,/\,\ensuremath{.375_{\pm\!.016}}} & \shortstack{\ensuremath{.576_{\pm\!.001}}\,/\,\ensuremath{.601_{\pm\!.025}}\\\ensuremath{.374_{\pm\!.019}}\,/\,\ensuremath{.374_{\pm\!.019}}}\\
& DCPO & \shortstack{\ensuremath{.490_{\pm\!.012}}\,/\,\ensuremath{.570_{\pm\!.038}}\\\ensuremath{.436_{\pm\!.032}}\,/\,\ensuremath{.435_{\pm\!.032}}} & \shortstack{\ensuremath{.660_{\pm\!.034}}\,/\,\ensuremath{.629_{\pm\!.007}}\\\ensuremath{.347_{\pm\!.016}}\,/\,\ensuremath{.347_{\pm\!.016}}} & \shortstack{\ensuremath{.850_{\pm\!.008}}\,/\,\ensuremath{.565_{\pm\!.017}}\\\ensuremath{.251_{\pm\!.012}}\,/\,\ensuremath{.251_{\pm\!.012}}} & \shortstack{\ensuremath{.604_{\pm\!.039}}\,/\,\ensuremath{.507_{\pm\!.013}}\\\ensuremath{.393_{\pm\!.043}}\,/\,\ensuremath{.392_{\pm\!.045}}} & \shortstack{\ensuremath{.561_{\pm\!.021}}\,/\,\ensuremath{.621_{\pm\!.013}}\\\ensuremath{.347_{\pm\!.022}}\,/\,\ensuremath{.347_{\pm\!.023}}}\\
& VL-Cal. & \shortstack{\ensuremath{.468_{\pm\!.004}}\,/\,\ensuremath{.643_{\pm\!.044}}\\\ensuremath{.271_{\pm\!.065}}\,/\,\ensuremath{\mathbf{.308}_{\pm\!.041}}} & \shortstack{\ensuremath{.639_{\pm\!.053}}\,/\,\ensuremath{.665_{\pm\!.065}}\\\ensuremath{.355_{\pm\!.071}}\,/\,\ensuremath{.361_{\pm\!.071}}} & \shortstack{\ensuremath{.816_{\pm\!.010}}\,/\,\ensuremath{.661_{\pm\!.047}}\\\ensuremath{.330_{\pm\!.019}}\,/\,\ensuremath{.328_{\pm\!.033}}} & \shortstack{\ensuremath{.615_{\pm\!.018}}\,/\,\ensuremath{.532_{\pm\!.049}}\\\ensuremath{.266_{\pm\!.117}}\,/\,\ensuremath{.308_{\pm\!.070}}} & \shortstack{\ensuremath{.560_{\pm\!.005}}\,/\,\ensuremath{.661_{\pm\!.025}}\\\ensuremath{.264_{\pm\!.056}}\,/\,\ensuremath{\mathbf{.292}_{\pm\!.039}}}\\
& GRPO (RLVR) & \shortstack{\ensuremath{.493_{\pm\!.005}}\,/\,\ensuremath{.501_{\pm\!.002}}\\\ensuremath{.320_{\pm\!.238}}\,/\,\ensuremath{.390_{\pm\!.128}}} & \shortstack{\ensuremath{.703_{\pm\!.020}}\,/\,\ensuremath{.502_{\pm\!.003}}\\\ensuremath{.234_{\pm\!.064}}\,/\,\ensuremath{.265_{\pm\!.037}}} & \shortstack{\ensuremath{.850_{\pm\!.002}}\,/\,\ensuremath{.500_{\pm\!.000}}\\\ensuremath{.171_{\pm\!.132}}\,/\,\ensuremath{.168_{\pm\!.049}}} & \shortstack{\ensuremath{.620_{\pm\!.039}}\,/\,\ensuremath{.505_{\pm\!.008}}\\\ensuremath{.219_{\pm\!.160}}\,/\,\ensuremath{.300_{\pm\!.050}}} & \shortstack{\ensuremath{.586_{\pm\!.005}}\,/\,\ensuremath{.502_{\pm\!.003}}\\\ensuremath{.286_{\pm\!.138}}\,/\,\ensuremath{.331_{\pm\!.082}}}\\
& P(True) & \shortstack{\ensuremath{.493_{\pm\!.005}}\,/\,\ensuremath{.589_{\pm\!.004}}\\\ensuremath{.499_{\pm\!.007}}\,/\,\ensuremath{.498_{\pm\!.007}}} & \shortstack{\ensuremath{.703_{\pm\!.020}}\,/\,\ensuremath{\mathbf{.738}_{\pm\!.017}}\\\ensuremath{.283_{\pm\!.020}}\,/\,\ensuremath{.280_{\pm\!.020}}} & \shortstack{\ensuremath{.850_{\pm\!.002}}\,/\,\ensuremath{\mathbf{.784}_{\pm\!.012}}\\\ensuremath{.147_{\pm\!.003}}\,/\,\ensuremath{.146_{\pm\!.004}}} & \shortstack{\ensuremath{.620_{\pm\!.039}}\,/\,\ensuremath{.453_{\pm\!.073}}\\\ensuremath{.459_{\pm\!.083}}\,/\,\ensuremath{.446_{\pm\!.082}}} & \shortstack{\ensuremath{.586_{\pm\!.005}}\,/\,\ensuremath{.594_{\pm\!.015}}\\\ensuremath{.420_{\pm\!.016}}\,/\,\ensuremath{.417_{\pm\!.015}}}\\
& Self-prob. & \shortstack{\ensuremath{.493_{\pm\!.005}}\,/\,\ensuremath{.552_{\pm\!.040}}\\\ensuremath{.422_{\pm\!.026}}\,/\,\ensuremath{.441_{\pm\!.025}}} & \shortstack{\ensuremath{.703_{\pm\!.020}}\,/\,\ensuremath{.560_{\pm\!.049}}\\\ensuremath{.229_{\pm\!.064}}\,/\,\ensuremath{.263_{\pm\!.043}}} & \shortstack{\ensuremath{.850_{\pm\!.002}}\,/\,\ensuremath{.505_{\pm\!.038}}\\\ensuremath{.154_{\pm\!.058}}\,/\,\ensuremath{.165_{\pm\!.021}}} & \shortstack{\ensuremath{.620_{\pm\!.039}}\,/\,\ensuremath{.548_{\pm\!.094}}\\\ensuremath{.347_{\pm\!.027}}\,/\,\ensuremath{.351_{\pm\!.026}}} & \shortstack{\ensuremath{.586_{\pm\!.005}}\,/\,\ensuremath{.527_{\pm\!.045}}\\\ensuremath{.362_{\pm\!.034}}\,/\,\ensuremath{.376_{\pm\!.026}}}\\
& Prompt ens. & \shortstack{\ensuremath{.493_{\pm\!.005}}\,/\,\ensuremath{.539_{\pm\!.034}}\\\ensuremath{.399_{\pm\!.034}}\,/\,\ensuremath{.421_{\pm\!.021}}} & \shortstack{\ensuremath{.703_{\pm\!.020}}\,/\,\ensuremath{.559_{\pm\!.049}}\\\ensuremath{\mathbf{.218}_{\pm\!.048}}\,/\,\ensuremath{\mathbf{.252}_{\pm\!.032}}} & \shortstack{\ensuremath{.850_{\pm\!.002}}\,/\,\ensuremath{.571_{\pm\!.061}}\\\ensuremath{\mathbf{.105}_{\pm\!.026}}\,/\,\ensuremath{\mathbf{.137}_{\pm\!.008}}} & \shortstack{\ensuremath{.620_{\pm\!.039}}\,/\,\ensuremath{.545_{\pm\!.078}}\\\ensuremath{.314_{\pm\!.012}}\,/\,\ensuremath{.329_{\pm\!.012}}} & \shortstack{\ensuremath{.586_{\pm\!.005}}\,/\,\ensuremath{.537_{\pm\!.031}}\\\ensuremath{.332_{\pm\!.027}}\,/\,\ensuremath{.351_{\pm\!.023}}}\\
& SAPLMA & \shortstack{\ensuremath{.493_{\pm\!.005}}\,/\,\ensuremath{.651_{\pm\!.023}}\\\ensuremath{.274_{\pm\!.062}}\,/\,\ensuremath{.315_{\pm\!.042}}} & \shortstack{\ensuremath{.703_{\pm\!.020}}\,/\,\ensuremath{.689_{\pm\!.065}}\\\ensuremath{.352_{\pm\!.203}}\,/\,\ensuremath{.362_{\pm\!.167}}} & \shortstack{\ensuremath{.850_{\pm\!.002}}\,/\,\ensuremath{.651_{\pm\!.049}}\\\ensuremath{.410_{\pm\!.262}}\,/\,\ensuremath{.384_{\pm\!.231}}} & \shortstack{\ensuremath{.620_{\pm\!.039}}\,/\,\ensuremath{\mathbf{.653}_{\pm\!.058}}\\\ensuremath{.234_{\pm\!.066}}\,/\,\ensuremath{.269_{\pm\!.035}}} & \shortstack{\ensuremath{.586_{\pm\!.005}}\,/\,\ensuremath{\mathbf{.687}_{\pm\!.018}}\\\ensuremath{.262_{\pm\!.084}}\,/\,\ensuremath{.299_{\pm\!.068}}}\\
& BICR & \shortstack{\ensuremath{.493_{\pm\!.005}}\,/\,\ensuremath{.587_{\pm\!.026}}\\\ensuremath{\mathbf{.228}_{\pm\!.072}}\,/\,\ensuremath{\mathbf{.308}_{\pm\!.038}}} & \shortstack{\ensuremath{.703_{\pm\!.020}}\,/\,\ensuremath{.580_{\pm\!.054}}\\\ensuremath{.262_{\pm\!.118}}\,/\,\ensuremath{.305_{\pm\!.074}}} & \shortstack{\ensuremath{.850_{\pm\!.002}}\,/\,\ensuremath{.552_{\pm\!.103}}\\\ensuremath{.411_{\pm\!.270}}\,/\,\ensuremath{.371_{\pm\!.200}}} & \shortstack{\ensuremath{.620_{\pm\!.039}}\,/\,\ensuremath{.649_{\pm\!.051}}\\\ensuremath{.299_{\pm\!.073}}\,/\,\ensuremath{.313_{\pm\!.061}}} & \shortstack{\ensuremath{.586_{\pm\!.005}}\,/\,\ensuremath{.635_{\pm\!.022}}\\\ensuremath{.274_{\pm\!.045}}\,/\,\ensuremath{.304_{\pm\!.028}}}\\
& DualRead & \shortstack{\ensuremath{.493_{\pm\!.005}}\,/\,\ensuremath{.619_{\pm\!.025}}\\\ensuremath{.244_{\pm\!.118}}\,/\,\ensuremath{.312_{\pm\!.064}}} & \shortstack{\ensuremath{.703_{\pm\!.020}}\,/\,\ensuremath{.535_{\pm\!.063}}\\\ensuremath{.376_{\pm\!.175}}\,/\,\ensuremath{.395_{\pm\!.137}}} & \shortstack{\ensuremath{.850_{\pm\!.002}}\,/\,\ensuremath{.546_{\pm\!.061}}\\\ensuremath{.445_{\pm\!.259}}\,/\,\ensuremath{.393_{\pm\!.207}}} & \shortstack{\ensuremath{.620_{\pm\!.039}}\,/\,\ensuremath{.623_{\pm\!.076}}\\\ensuremath{\mathbf{.187}_{\pm\!.030}}\,/\,\ensuremath{\mathbf{.251}_{\pm\!.017}}} & \shortstack{\ensuremath{.586_{\pm\!.005}}\,/\,\ensuremath{.647_{\pm\!.034}}\\\ensuremath{\mathbf{.256}_{\pm\!.075}}\,/\,\ensuremath{.294_{\pm\!.059}}}\\
\bottomrule
\end{tabular}
\caption{Reliability on PMC-VQA, OmniMedVQA, ReXVQA, VQA-Med 2019, and OOD6; see Table~\ref{tab:ood_acc} for format.}
\label{tab:calib}
\end{table*}
\subsection{Per-domain OOD reliability diagrams}
\label{app:ood_reliability_diagrams}

\begin{figure*}[!tp]
\centering
\includegraphics[width=.235\textwidth]{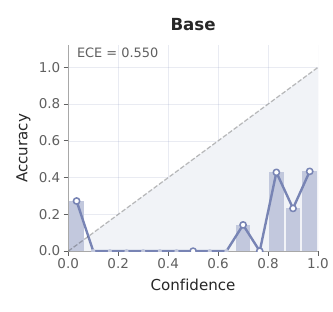}\hfill
\includegraphics[width=.235\textwidth]{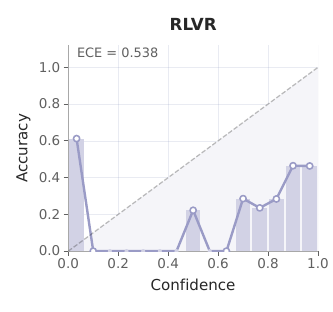}\hfill
\includegraphics[width=.235\textwidth]{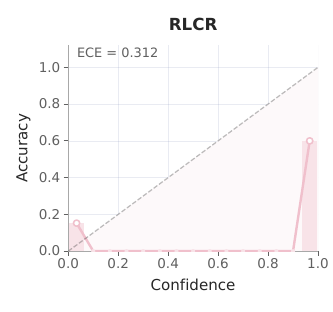}\hfill
\includegraphics[width=.235\textwidth]{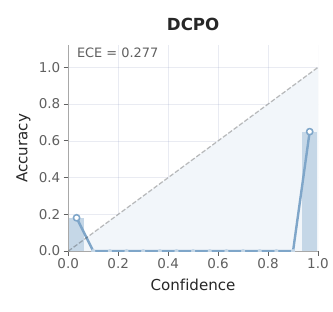}\\[2pt]
\includegraphics[width=.235\textwidth]{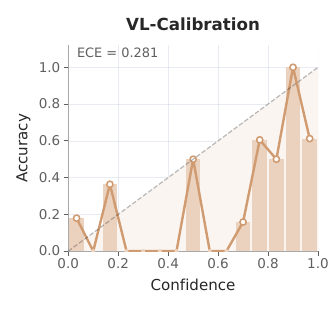}\hfill
\includegraphics[width=.235\textwidth]{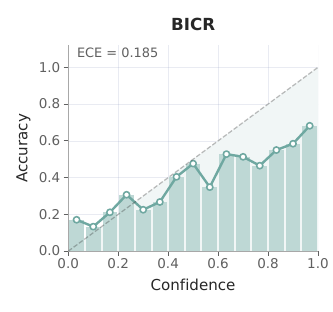}\hfill
\includegraphics[width=.235\textwidth]{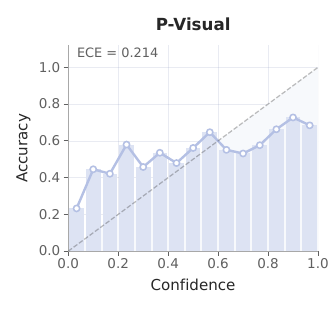}\hfill
\includegraphics[width=.235\textwidth]{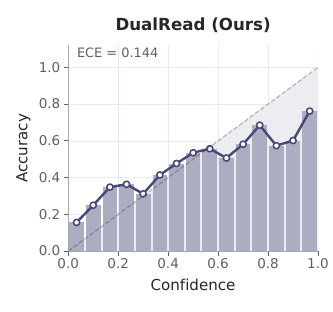}\\[2pt]
\includegraphics[width=.235\textwidth]{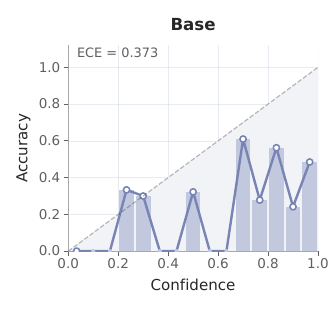}\hfill
\includegraphics[width=.235\textwidth]{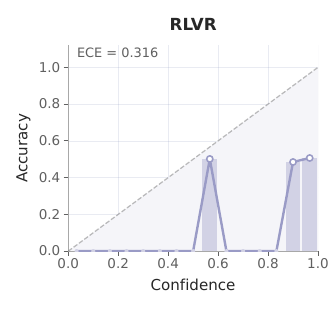}\hfill
\includegraphics[width=.235\textwidth]{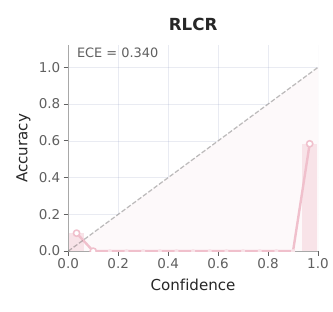}\hfill
\includegraphics[width=.235\textwidth]{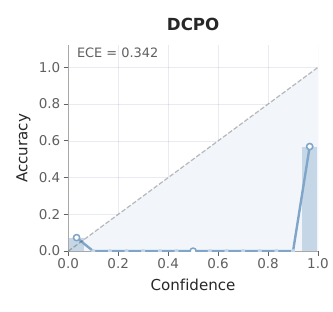}\\[2pt]
\includegraphics[width=.235\textwidth]{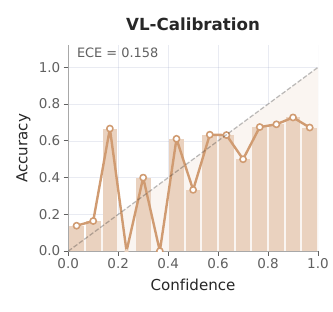}\hfill
\includegraphics[width=.235\textwidth]{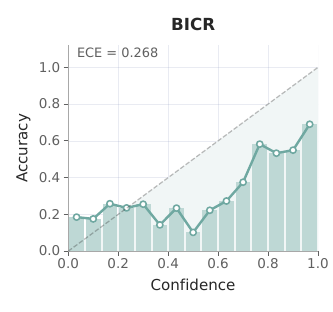}\hfill
\includegraphics[width=.235\textwidth]{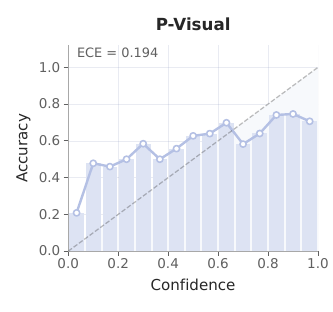}\hfill
\includegraphics[width=.235\textwidth]{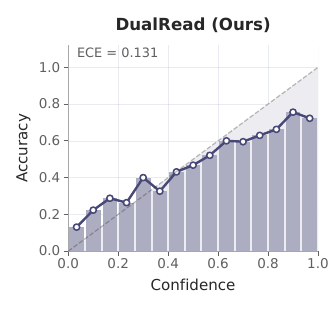}
\caption{VQA-RAD reliability ($n=600$; short answer). Qwen rows precede MedGemma rows.}
\label{fig:ood_rel_vqarad}
\end{figure*}

\begin{figure*}[!tp]
\centering
\includegraphics[width=.235\textwidth]{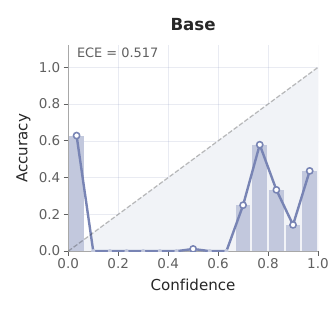}\hfill
\includegraphics[width=.235\textwidth]{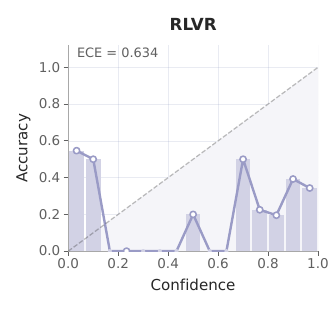}\hfill
\includegraphics[width=.235\textwidth]{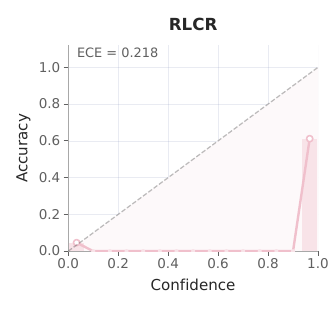}\hfill
\includegraphics[width=.235\textwidth]{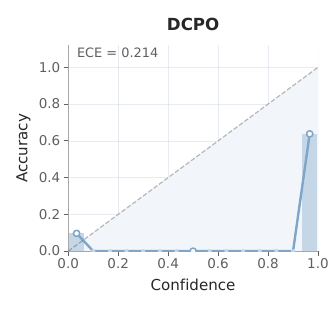}\\[2pt]
\includegraphics[width=.235\textwidth]{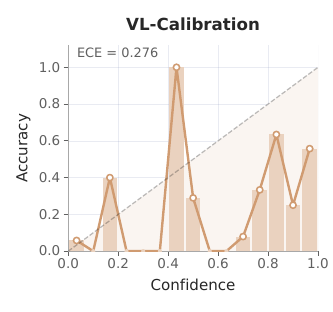}\hfill
\includegraphics[width=.235\textwidth]{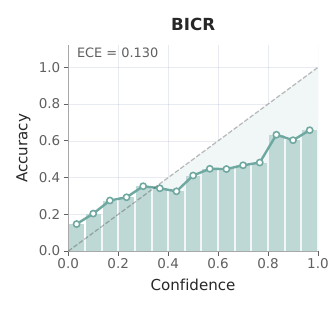}\hfill
\includegraphics[width=.235\textwidth]{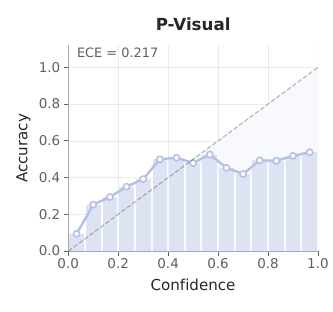}\hfill
\includegraphics[width=.235\textwidth]{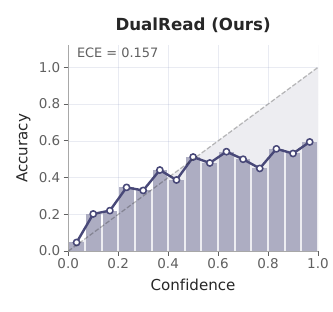}\\[2pt]
\includegraphics[width=.235\textwidth]{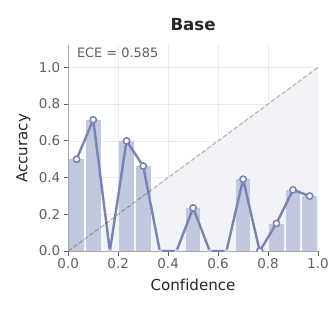}\hfill
\includegraphics[width=.235\textwidth]{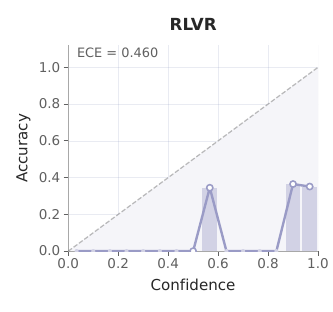}\hfill
\includegraphics[width=.235\textwidth]{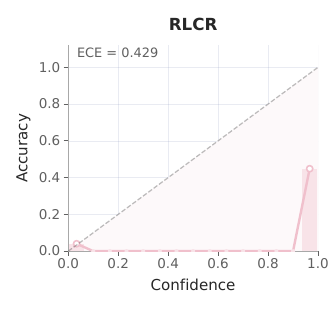}\hfill
\includegraphics[width=.235\textwidth]{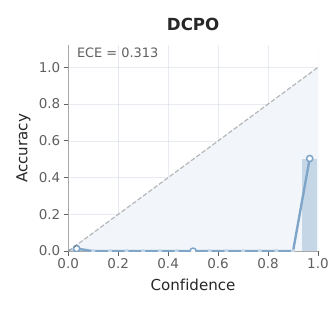}\\[2pt]
\includegraphics[width=.235\textwidth]{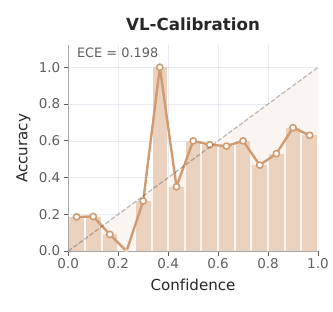}\hfill
\includegraphics[width=.235\textwidth]{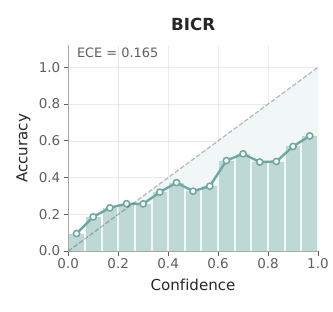}\hfill
\includegraphics[width=.235\textwidth]{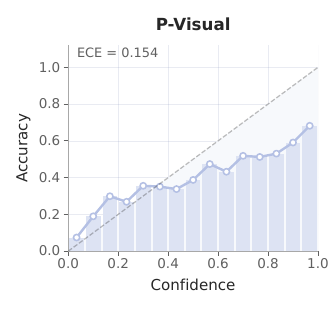}\hfill
\includegraphics[width=.235\textwidth]{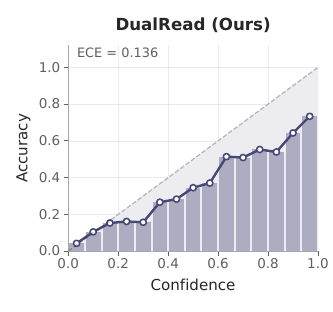}
\caption{PathVQA reliability ($n=1000$; short answer). Qwen rows precede MedGemma rows.}
\label{fig:ood_rel_pathvqa}
\end{figure*}

\begin{figure*}[!tp]
\centering
\includegraphics[width=.235\textwidth]{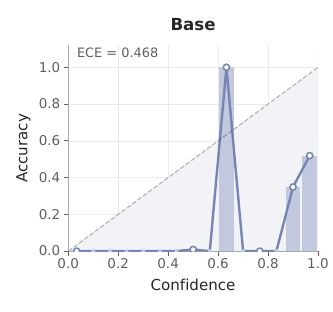}\hfill
\includegraphics[width=.235\textwidth]{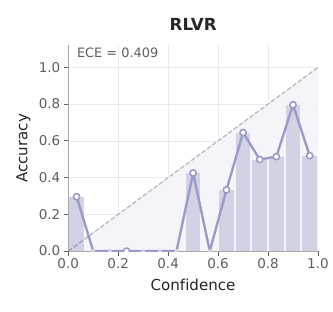}\hfill
\includegraphics[width=.235\textwidth]{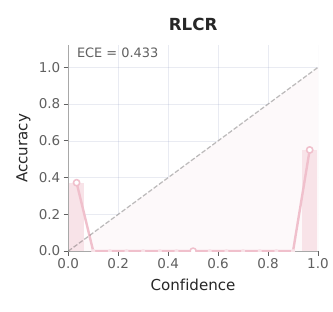}\hfill
\includegraphics[width=.235\textwidth]{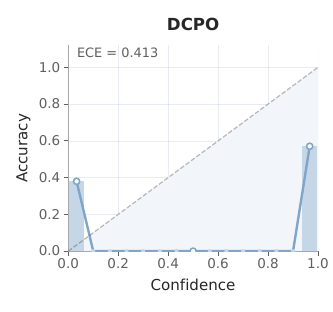}\\[2pt]
\includegraphics[width=.235\textwidth]{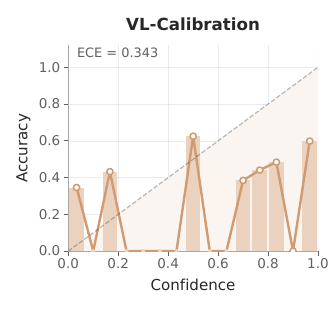}\hfill
\includegraphics[width=.235\textwidth]{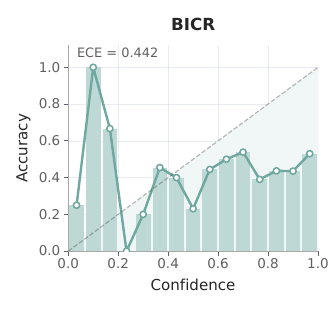}\hfill
\includegraphics[width=.235\textwidth]{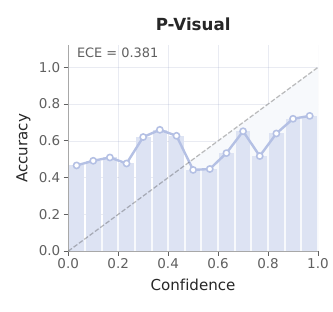}\hfill
\includegraphics[width=.235\textwidth]{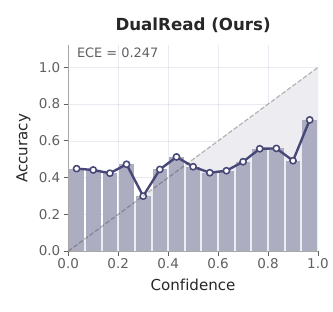}\\[2pt]
\includegraphics[width=.235\textwidth]{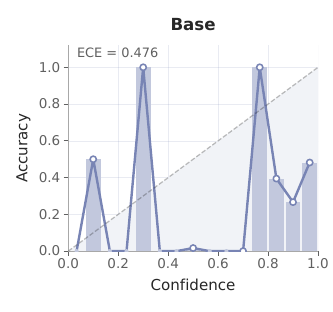}\hfill
\includegraphics[width=.235\textwidth]{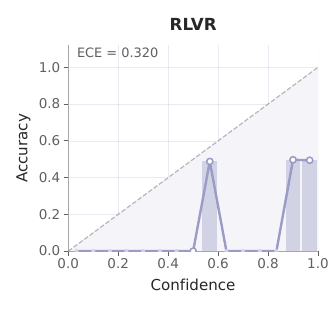}\hfill
\includegraphics[width=.235\textwidth]{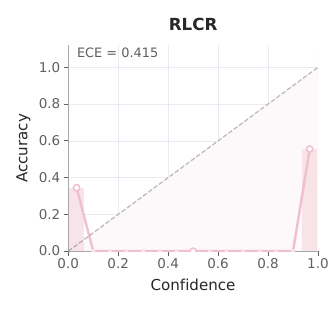}\hfill
\includegraphics[width=.235\textwidth]{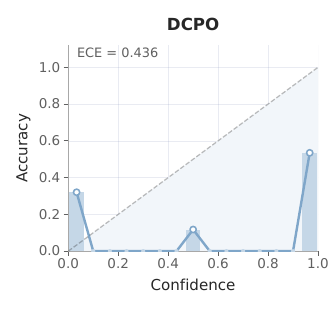}\\[2pt]
\includegraphics[width=.235\textwidth]{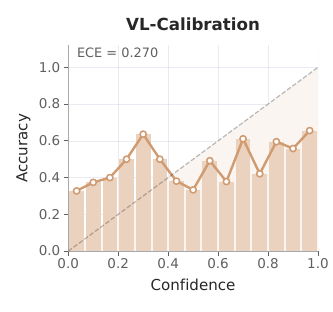}\hfill
\includegraphics[width=.235\textwidth]{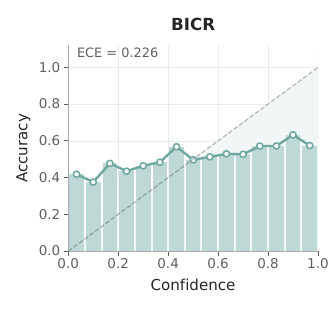}\hfill
\includegraphics[width=.235\textwidth]{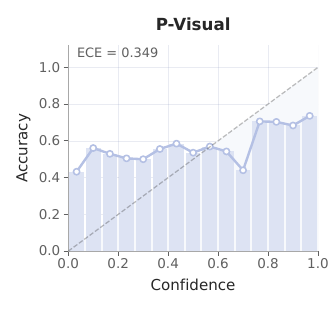}\hfill
\includegraphics[width=.235\textwidth]{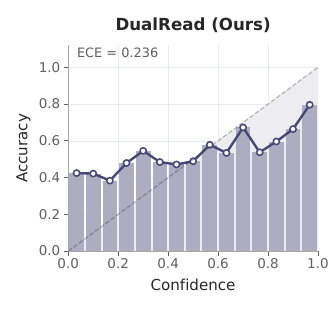}
\caption{PMC-VQA reliability ($n=1000$; option visible). Qwen rows precede MedGemma rows.}
\label{fig:ood_rel_pmcvqa}
\end{figure*}

\begin{figure*}[!tp]
\centering
\includegraphics[width=.235\textwidth]{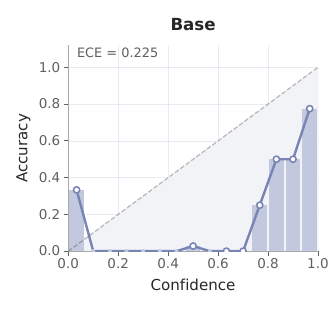}\hfill
\includegraphics[width=.235\textwidth]{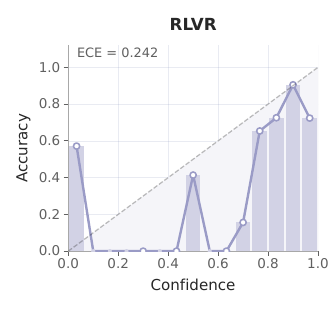}\hfill
\includegraphics[width=.235\textwidth]{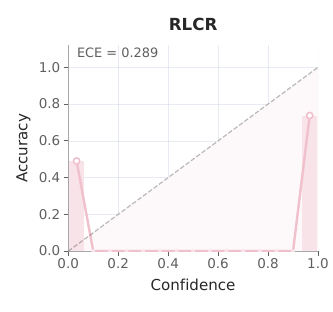}\hfill
\includegraphics[width=.235\textwidth]{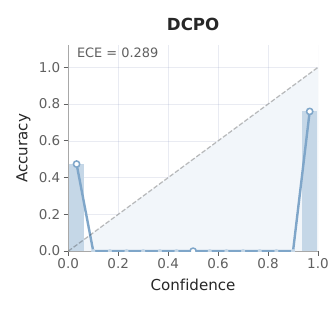}\\[2pt]
\includegraphics[width=.235\textwidth]{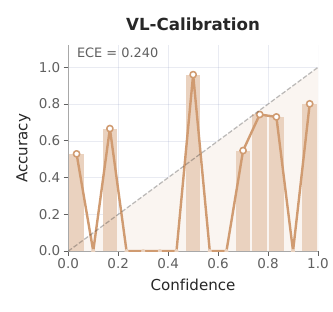}\hfill
\includegraphics[width=.235\textwidth]{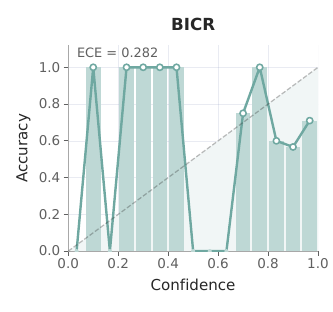}\hfill
\includegraphics[width=.235\textwidth]{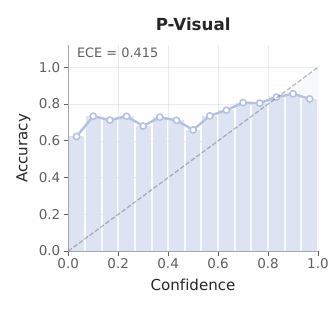}\hfill
\includegraphics[width=.235\textwidth]{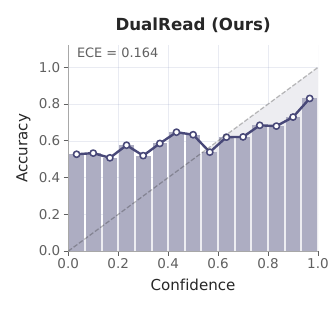}\\[2pt]
\includegraphics[width=.235\textwidth]{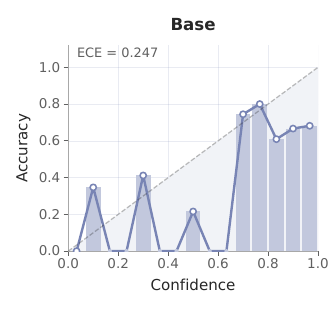}\hfill
\includegraphics[width=.235\textwidth]{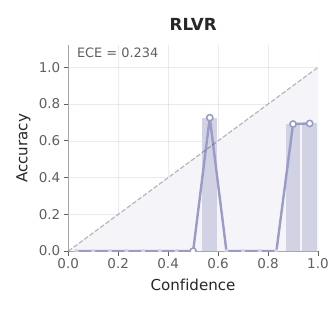}\hfill
\includegraphics[width=.235\textwidth]{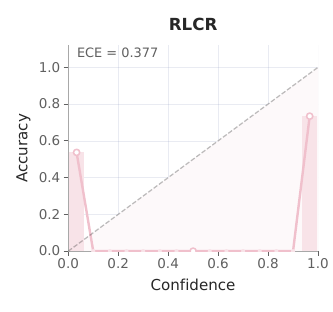}\hfill
\includegraphics[width=.235\textwidth]{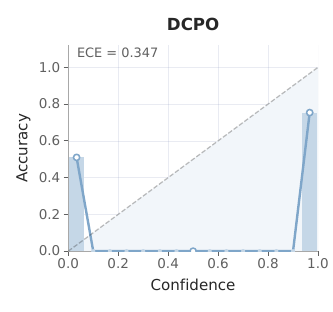}\\[2pt]
\includegraphics[width=.235\textwidth]{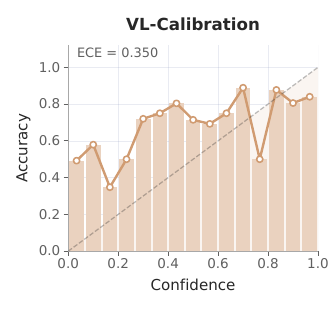}\hfill
\includegraphics[width=.235\textwidth]{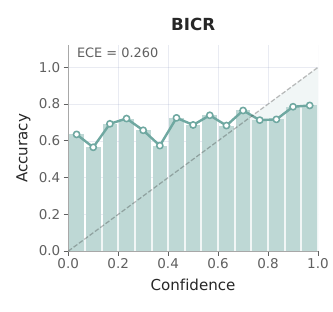}\hfill
\includegraphics[width=.235\textwidth]{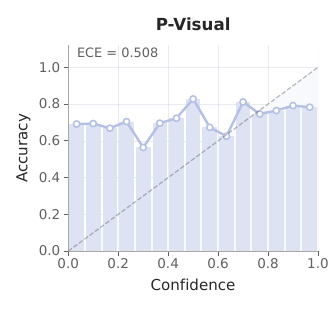}\hfill
\includegraphics[width=.235\textwidth]{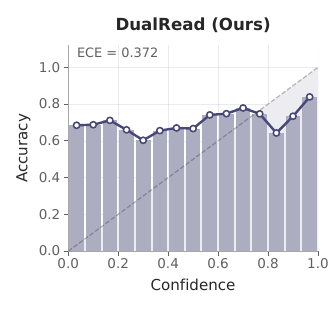}
\caption{OmniMedVQA reliability ($n=920$; option visible). Qwen rows precede MedGemma rows.}
\label{fig:ood_rel_omnimedvqa}
\end{figure*}

\begin{figure*}[!tp]
\centering
\includegraphics[width=.235\textwidth]{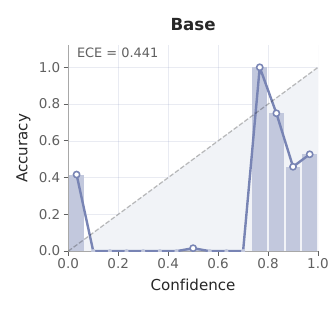}\hfill
\includegraphics[width=.235\textwidth]{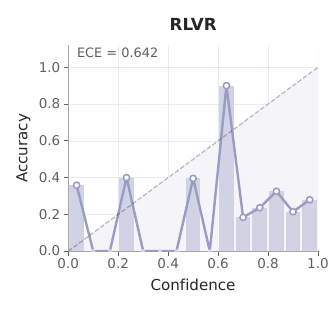}\hfill
\includegraphics[width=.235\textwidth]{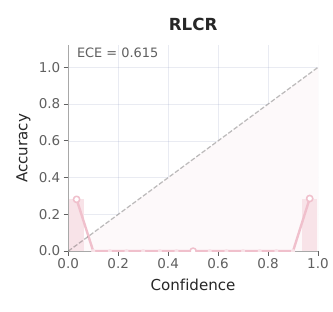}\hfill
\includegraphics[width=.235\textwidth]{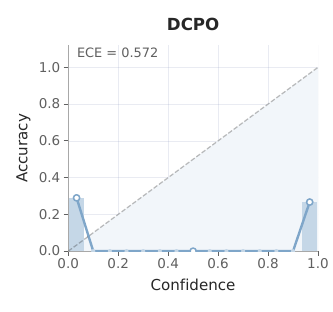}\\[2pt]
\includegraphics[width=.235\textwidth]{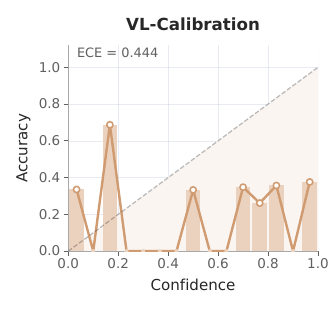}\hfill
\includegraphics[width=.235\textwidth]{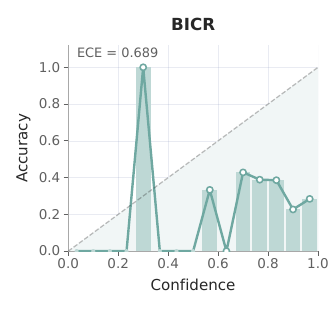}\hfill
\includegraphics[width=.235\textwidth]{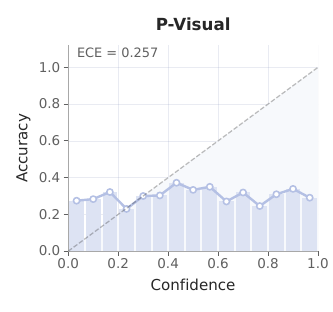}\hfill
\includegraphics[width=.235\textwidth]{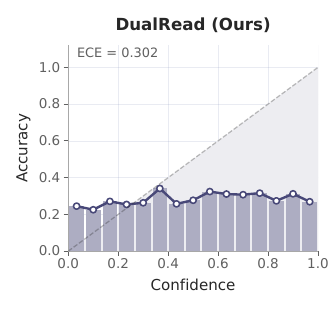}\\[2pt]
\includegraphics[width=.235\textwidth]{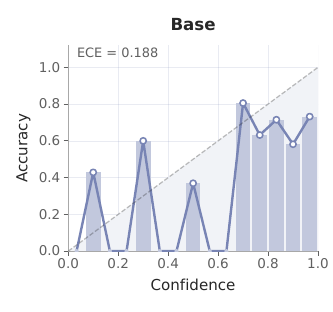}\hfill
\includegraphics[width=.235\textwidth]{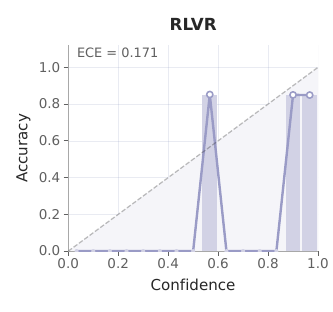}\hfill
\includegraphics[width=.235\textwidth]{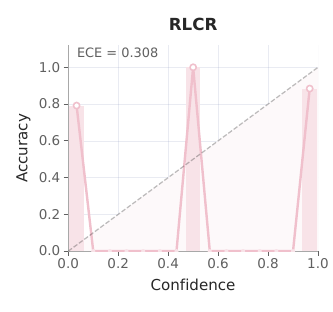}\hfill
\includegraphics[width=.235\textwidth]{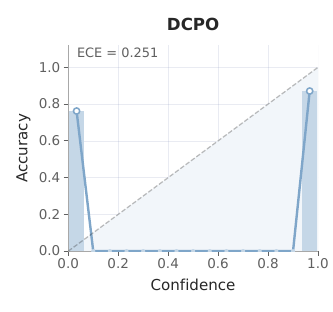}\\[2pt]
\includegraphics[width=.235\textwidth]{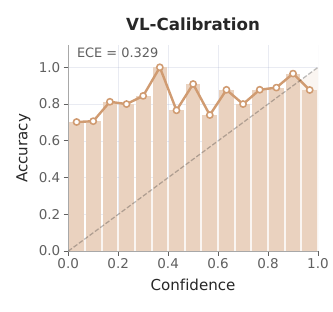}\hfill
\includegraphics[width=.235\textwidth]{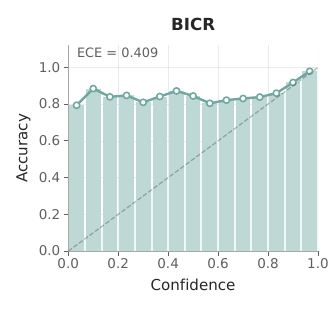}\hfill
\includegraphics[width=.235\textwidth]{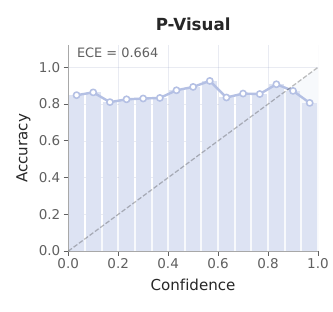}\hfill
\includegraphics[width=.235\textwidth]{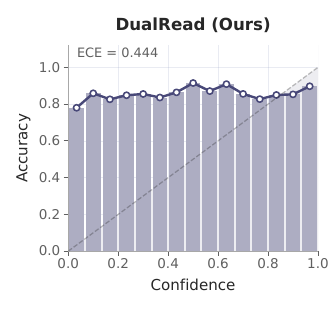}
\caption{ReXVQA reliability ($n=1000$; option visible). Qwen rows precede MedGemma rows.}
\label{fig:ood_rel_rexvqa}
\end{figure*}

\begin{figure*}[!tp]
\centering
\includegraphics[width=.235\textwidth]{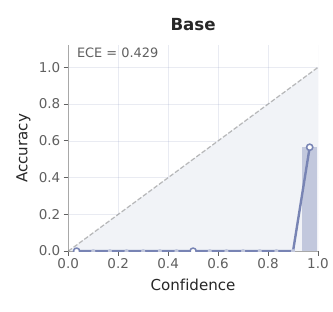}\hfill
\includegraphics[width=.235\textwidth]{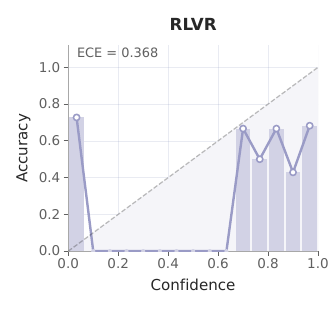}\hfill
\includegraphics[width=.235\textwidth]{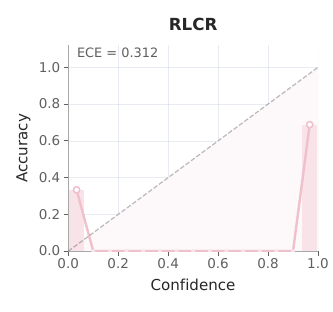}\hfill
\includegraphics[width=.235\textwidth]{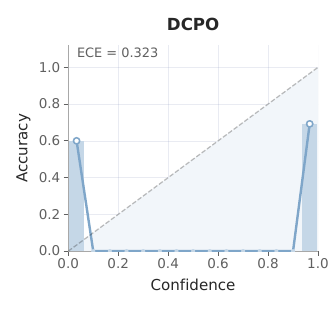}\\[2pt]
\includegraphics[width=.235\textwidth]{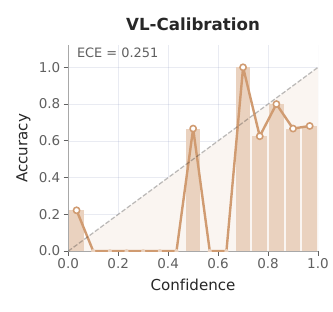}\hfill
\includegraphics[width=.235\textwidth]{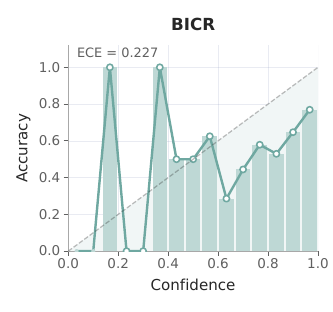}\hfill
\includegraphics[width=.235\textwidth]{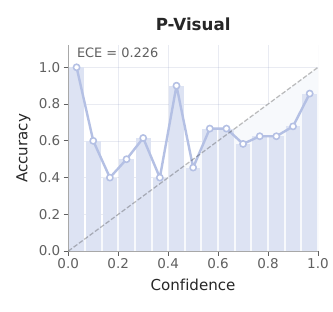}\hfill
\includegraphics[width=.235\textwidth]{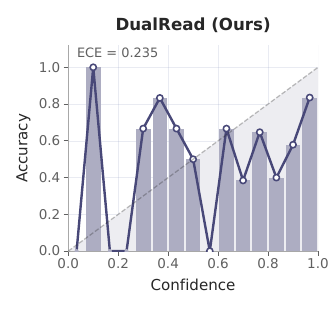}\\[2pt]
\includegraphics[width=.235\textwidth]{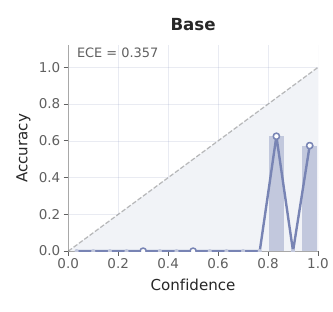}\hfill
\includegraphics[width=.235\textwidth]{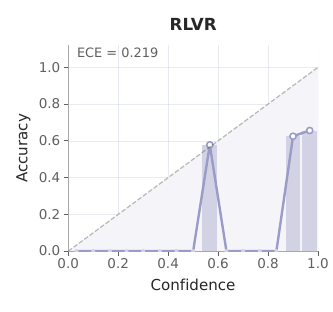}\hfill
\includegraphics[width=.235\textwidth]{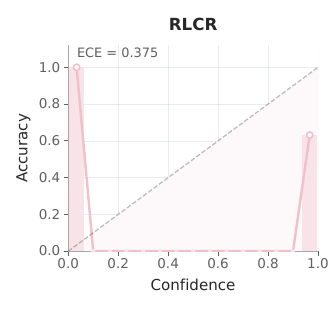}\hfill
\includegraphics[width=.235\textwidth]{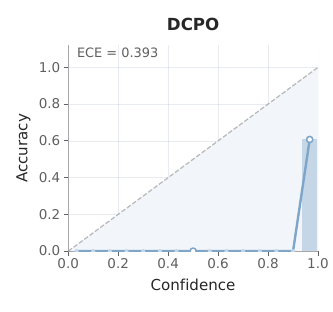}\\[2pt]
\includegraphics[width=.235\textwidth]{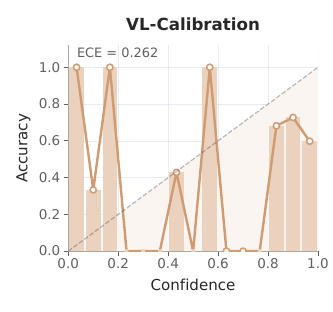}\hfill
\includegraphics[width=.235\textwidth]{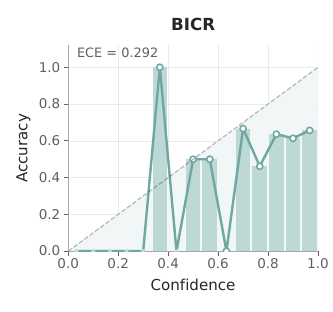}\hfill
\includegraphics[width=.235\textwidth]{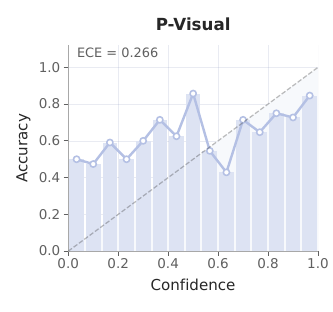}\hfill
\includegraphics[width=.235\textwidth]{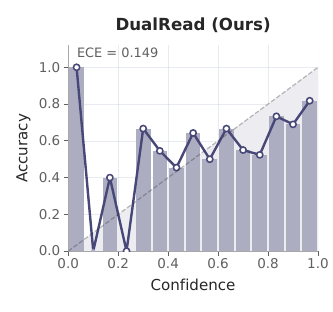}
\caption{VQA-Med 2019 reliability ($n=64$; closed answer). Qwen rows precede MedGemma rows.}
\label{fig:ood_rel_vqamed_closed}
\end{figure*}

Figures~\ref{fig:ood_rel_vqarad}--\ref{fig:ood_rel_vqamed_closed} show the
calibration behavior behind the six per-domain OOD rows above, for the complete
roster of eight confidence interfaces rather than a two-method excerpt.  Every
panel uses the exact frozen per-item predictions used to compute
Tables~\ref{tab:ood_acc} and~\ref{tab:calib}; no
OOD example is used to fit a reader, temperature, bin edge, or other parameter.
Each figure carries one domain, with the first two rows for Qwen3VL-2B and the
last two for MedGemma1.5-4B, and the same left-to-right method order in every
row pair.  The three equal-size seed arrays are concatenated only to make each
reliability diagram readable, so the ECE printed inside a panel is the pooled
diagnostic ECE, whereas the tables report the mean and sample standard
deviation of the three separately computed seed-level ECE values.  All panels
use 15 fixed equal-width bins on the full $[0,1]$ confidence range; empty bins
carry no mass and do not enter ECE.  Panels are drawn on a common axis so that
the occupied confidence support of each interface is directly comparable.

\paragraph{Confidence-support audit.}
Reliability diagrams bin the confidence axis and therefore do not show how many
distinct values a confidence channel emits.  Table~\ref{tab:channel}
records that quantity directly for the seed-43 SLAKE-test predictions.  Qwen RLVR
assigns confidence one to $1060/1061$ examples and MedGemma RLVR assigns one to all
$1061$ examples; RLCR emits two distinct endpoint values; DualRead emits a
distinct non-endpoint score for every example.  This audit measures score
resolution only, separately from how the scores rank correctness.
For this mechanism audit, support statistics are computed over valid parsed scores and
the valid rate is reported separately; the deployment metrics retain invalid examples
and impute confidence $0.5$ as specified in Section~\ref{app:evaluation_protocol}.

\paragraph{Answer-type reliability.}
Table~\ref{tab:reliability} compares open- and closed-question reliability and selective prediction.

\begin{table*}[!t]
\centering
\footnotesize
\setlength{\tabcolsep}{3pt}
\renewcommand{\arraystretch}{1.04}
\begin{tabular*}{\textwidth}{@{\extracolsep{\fill}}lllrrrrrr@{}}
\toprule
Backbone & Dataset & Method & Acc$\uparrow$ & AUROC$\uparrow$ & ECE$\downarrow$ & Brier$\downarrow$ & AURC$\downarrow$ & HCE@.9$\downarrow$\\
\midrule
\multicolumn{9}{@{}l}{\emph{Open questions}}\\
\multirow{4}{*}{Qwen3VL-2B}
& \multirow{2}{*}{SLAKE} & RLCR & $.480_{\pm.011}$ & $.885_{\pm.016}$ & $.117_{\pm.018}$ & $.117_{\pm.018}$ & $.261_{\pm.026}$ & $.083_{\pm.023}$\\
& & DualRead & $.518_{\pm.004}$ & $\mathbf{.973}_{\pm.006}$ & $\mathbf{.039}_{\pm.001}$ & $\mathbf{.063}_{\pm.005}$ & $\mathbf{.155}_{\pm.003}$ & $\mathbf{.005}_{\pm.003}$\\
\addlinespace[1pt]
& \multirow{2}{*}{VQA-RAD} & RLCR & $.143_{\pm.009}$ & $\mathbf{.630}_{\pm.013}$ & $\mathbf{.210}_{\pm.016}$ & $.210_{\pm.016}$ & $.792_{\pm.013}$ & $.126_{\pm.027}$\\
& & DualRead & $.158_{\pm.005}$ & $.554_{\pm.047}$ & $.223_{\pm.029}$ & $\mathbf{.205}_{\pm.025}$ & $\mathbf{.775}_{\pm.018}$ & $\mathbf{.071}_{\pm.009}$\\
\addlinespace[2pt]
\multirow{4}{*}{MedGemma1.5-4B}
& \multirow{2}{*}{SLAKE} & RLCR & $.552_{\pm.034}$ & $.828_{\pm.024}$ & $.164_{\pm.024}$ & $.164_{\pm.024}$ & $.247_{\pm.046}$ & $.124_{\pm.041}$\\
& & DualRead & $.581_{\pm.023}$ & $\mathbf{.970}_{\pm.005}$ & $\mathbf{.053}_{\pm.015}$ & $\mathbf{.069}_{\pm.003}$ & $\mathbf{.117}_{\pm.013}$ & $\mathbf{.003}_{\pm.002}$\\
\addlinespace[1pt]
& \multirow{2}{*}{VQA-RAD} & RLCR & $.143_{\pm.025}$ & $.635_{\pm.093}$ & $.394_{\pm.092}$ & $.394_{\pm.092}$ & $.811_{\pm.004}$ & $.346_{\pm.081}$\\
& & DualRead & $.151_{\pm.017}$ & $\mathbf{.640}_{\pm.028}$ & $\mathbf{.209}_{\pm.023}$ & $\mathbf{.202}_{\pm.038}$ & $\mathbf{.758}_{\pm.021}$ & $\mathbf{.072}_{\pm.039}$\\
\midrule
\multicolumn{9}{@{}l}{\emph{Closed questions}}\\
\multirow{4}{*}{Qwen3VL-2B}
& \multirow{2}{*}{SLAKE} & RLCR & $.808_{\pm.006}$ & $.516_{\pm.008}$ & $.200_{\pm.007}$ & $.200_{\pm.007}$ & $.187_{\pm.009}$ & $.182_{\pm.010}$\\
& & DualRead & $.828_{\pm.018}$ & $\mathbf{.841}_{\pm.010}$ & $\mathbf{.065}_{\pm.004}$ & $\mathbf{.114}_{\pm.005}$ & $\mathbf{.047}_{\pm.005}$ & $\mathbf{.010}_{\pm.004}$\\
\addlinespace[1pt]
& \multirow{2}{*}{VQA-RAD} & RLCR & $.633_{\pm.011}$ & $.506_{\pm.011}$ & $.377_{\pm.014}$ & $.377_{\pm.014}$ & $.364_{\pm.013}$ & $.342_{\pm.017}$\\
& & DualRead & $.644_{\pm.007}$ & $\mathbf{.641}_{\pm.009}$ & $\mathbf{.174}_{\pm.016}$ & $\mathbf{.254}_{\pm.010}$ & $\mathbf{.235}_{\pm.014}$ & $\mathbf{.020}_{\pm.010}$\\
\addlinespace[2pt]
\multirow{4}{*}{MedGemma1.5-4B}
& \multirow{2}{*}{SLAKE} & RLCR & $.819_{\pm.017}$ & $.517_{\pm.016}$ & $.183_{\pm.014}$ & $.183_{\pm.014}$ & $.176_{\pm.015}$ & $.172_{\pm.015}$\\
& & DualRead & $.825_{\pm.009}$ & $\mathbf{.851}_{\pm.011}$ & $\mathbf{.044}_{\pm.017}$ & $\mathbf{.103}_{\pm.004}$ & $\mathbf{.049}_{\pm.003}$ & $\mathbf{.022}_{\pm.003}$\\
\addlinespace[1pt]
& \multirow{2}{*}{VQA-RAD} & RLCR & $.679_{\pm.049}$ & $.528_{\pm.039}$ & $.305_{\pm.017}$ & $.305_{\pm.017}$ & $.308_{\pm.030}$ & $.298_{\pm.019}$\\
& & DualRead & $.723_{\pm.015}$ & $\mathbf{.617}_{\pm.013}$ & $\mathbf{.151}_{\pm.014}$ & $\mathbf{.227}_{\pm.016}$ & $\mathbf{.201}_{\pm.007}$ & $\mathbf{.060}_{\pm.037}$\\
\bottomrule
\end{tabular*}
\caption{Open/closed reliability for RLCR and DualRead; bold excludes Acc.}
\label{tab:reliability}
\end{table*}

\section{Grounding Details}
\label{app:grounding}

Table~\ref{tab:ccg} consolidates per-domain OOD grounding, the paired SLAKE
actor/confidence audit, and decisive-set eligibility under the frozen CCG-AUC
protocol in Section~\ref{app:ccg_properties}. Confidence readers and
temperatures are not refit on these evaluation sets.

\begin{table*}[!t]
\centering
\footnotesize
\setlength{\tabcolsep}{3pt}
\renewcommand{\arraystretch}{1.02}
\textit{(a) Per-domain OOD CCG-AUC}\par\smallskip
\begin{tabular*}{\textwidth}{@{\extracolsep{\fill}}llcccccc@{}}
\toprule
& & \multicolumn{2}{c}{VQA-RAD} & \multicolumn{2}{c}{PathVQA} & \multicolumn{2}{c}{VQA-Med 2019 closed}\\
\cmidrule(lr){3-4}\cmidrule(lr){5-6}\cmidrule(lr){7-8}
Backbone & Method & Nat.$\uparrow$ & Strict$\uparrow$ & Nat.$\uparrow$ & Strict$\uparrow$ & Nat.$\uparrow$ & Strict$\uparrow$\\
\midrule
\multirow{11}{*}{\rotatebox[origin=c]{90}{Qwen3VL-2B}}
& Base & \ensuremath{\mathbf{.786}} & \ensuremath{.700} & \ensuremath{.692} & \ensuremath{.500} & \ensuremath{\mathbf{.867}} & \ensuremath{.667}\\
& RLCR & \ensuremath{.500_{\pm\!.000}} & \ensuremath{.500_{\pm\!.000}} & \ensuremath{.521_{\pm\!.043}} & \ensuremath{.499_{\pm\!.040}} & \ensuremath{.528_{\pm\!.048}} & \ensuremath{.528_{\pm\!.048}}\\
& DCPO & \ensuremath{.519_{\pm\!.032}} & \ensuremath{.533_{\pm\!.058}} & \ensuremath{.492_{\pm\!.013}} & \ensuremath{.487_{\pm\!.022}} & \ensuremath{.506_{\pm\!.092}} & \ensuremath{.506_{\pm\!.092}}\\
& VL-Cal. & \ensuremath{.495_{\pm\!.064}} & \ensuremath{.486_{\pm\!.105}} & \ensuremath{.608_{\pm\!.065}} & \ensuremath{.593_{\pm\!.085}} & \ensuremath{.595_{\pm\!.135}} & \ensuremath{.597_{\pm\!.134}}\\
& GRPO (RLVR) & \ensuremath{.521_{\pm\!.036}} & \ensuremath{.500_{\pm\!.000}} & \ensuremath{.567_{\pm\!.115}} & \ensuremath{.548_{\pm\!.082}} & \ensuremath{.500_{\pm\!.000}} & \ensuremath{.500_{\pm\!.000}}\\
& P(True) & \ensuremath{.419_{\pm\!.038}} & \ensuremath{.250_{\pm\!.000}} & \ensuremath{.692_{\pm\!.095}} & \ensuremath{.556_{\pm\!.076}} & \ensuremath{.431_{\pm\!.237}} & \ensuremath{.431_{\pm\!.237}}\\
& Self-prob. & \ensuremath{.421_{\pm\!.028}} & \ensuremath{.292_{\pm\!.144}} & \ensuremath{.518_{\pm\!.055}} & \ensuremath{.500_{\pm\!.071}} & \ensuremath{.535_{\pm\!.107}} & \ensuremath{.535_{\pm\!.107}}\\
& Prompt ens. & \ensuremath{.440_{\pm\!.004}} & \ensuremath{.292_{\pm\!.072}} & \ensuremath{.560_{\pm\!.104}} & \ensuremath{.512_{\pm\!.055}} & \ensuremath{.556_{\pm\!.136}} & \ensuremath{.556_{\pm\!.136}}\\
& SAPLMA & \ensuremath{.509_{\pm\!.264}} & \ensuremath{.333_{\pm\!.382}} & \ensuremath{\mathbf{.709}_{\pm\!.010}} & \ensuremath{\mathbf{.687}_{\pm\!.048}} & \ensuremath{.431_{\pm\!.214}} & \ensuremath{.431_{\pm\!.214}}\\
& BICR & \ensuremath{.593_{\pm\!.160}} & \ensuremath{\mathbf{.833}_{\pm\!.144}} & \ensuremath{.455_{\pm\!.161}} & \ensuremath{.415_{\pm\!.137}} & \ensuremath{.542_{\pm\!.072}} & \ensuremath{.542_{\pm\!.072}}\\
& DualRead & \ensuremath{.556_{\pm\!.096}} & \ensuremath{.417_{\pm\!.144}} & \ensuremath{.511_{\pm\!.140}} & \ensuremath{.461_{\pm\!.091}} & \ensuremath{.806_{\pm\!.173}} & \ensuremath{\mathbf{.806}_{\pm\!.173}}\\
\midrule
\multirow{11}{*}{\rotatebox[origin=c]{90}{MedGemma1.5-4B}}
& Base & \ensuremath{.429} & \ensuremath{.375} & \ensuremath{.542} & \ensuremath{.250} & \ensuremath{.536} & \ensuremath{.500}\\
& RLCR & \ensuremath{.500_{\pm\!.000}} & \ensuremath{.500_{\pm\!.000}} & \ensuremath{.557_{\pm\!.013}} & \ensuremath{.486_{\pm\!.024}} & \ensuremath{.500_{\pm\!.000}} & \ensuremath{.500_{\pm\!.000}}\\
& DCPO & \ensuremath{.500_{\pm\!.000}} & \ensuremath{.500_{\pm\!.000}} & \ensuremath{\mathbf{.705}_{\pm\!.036}} & \ensuremath{\mathbf{.671}_{\pm\!.049}} & \ensuremath{.500_{\pm\!.000}} & \ensuremath{.500_{\pm\!.000}}\\
& VL-Cal. & \ensuremath{.524_{\pm\!.041}} & \ensuremath{.542_{\pm\!.072}} & \ensuremath{.675_{\pm\!.027}} & \ensuremath{.597_{\pm\!.020}} & \ensuremath{.596_{\pm\!.084}} & \ensuremath{.524_{\pm\!.100}}\\
& GRPO (RLVR) & \ensuremath{.500_{\pm\!.000}} & \ensuremath{.500_{\pm\!.000}} & \ensuremath{.511_{\pm\!.019}} & \ensuremath{.519_{\pm\!.032}} & \ensuremath{.500_{\pm\!.000}} & \ensuremath{.500_{\pm\!.000}}\\
& P(True) & \ensuremath{.387_{\pm\!.049}} & \ensuremath{.244_{\pm\!.214}} & \ensuremath{.605_{\pm\!.071}} & \ensuremath{.558_{\pm\!.128}} & \ensuremath{.623_{\pm\!.167}} & \ensuremath{.623_{\pm\!.167}}\\
& Self-prob. & \ensuremath{.493_{\pm\!.061}} & \ensuremath{.500_{\pm\!.000}} & \ensuremath{.606_{\pm\!.108}} & \ensuremath{.636_{\pm\!.127}} & \ensuremath{.576_{\pm\!.084}} & \ensuremath{.576_{\pm\!.084}}\\
& Prompt ens. & \ensuremath{.491_{\pm\!.159}} & \ensuremath{.411_{\pm\!.084}} & \ensuremath{.575_{\pm\!.206}} & \ensuremath{.574_{\pm\!.247}} & \ensuremath{.550_{\pm\!.109}} & \ensuremath{.550_{\pm\!.109}}\\
& SAPLMA & \ensuremath{.381_{\pm\!.206}} & \ensuremath{.644_{\pm\!.336}} & \ensuremath{.667_{\pm\!.185}} & \ensuremath{.661_{\pm\!.179}} & \ensuremath{.602_{\pm\!.040}} & \ensuremath{.602_{\pm\!.040}}\\
& BICR & \ensuremath{\mathbf{.646}_{\pm\!.067}} & \ensuremath{.656_{\pm\!.150}} & \ensuremath{.541_{\pm\!.096}} & \ensuremath{.574_{\pm\!.085}} & \ensuremath{.701_{\pm\!.221}} & \ensuremath{.701_{\pm\!.221}}\\
& DualRead & \ensuremath{.595_{\pm\!.367}} & \ensuremath{\mathbf{.756}_{\pm\!.214}} & \ensuremath{.540_{\pm\!.062}} & \ensuremath{.524_{\pm\!.099}} & \ensuremath{\mathbf{.769}_{\pm\!.184}} & \ensuremath{\mathbf{.769}_{\pm\!.184}}\\
\bottomrule
\end{tabular*}

\vspace{4pt}
\textit{(b) SLAKE actor audit and paired confidence comparison}\par\smallskip
\begin{tabular*}{\textwidth}{@{\extracolsep{\fill}}lrrrrrrrrr@{}}
\toprule
Backbone & Acc.$^{\rm real}$ & Acc.$^{\rm hn}$ & VRS & $|\mathcal E|$ & Src/Dnr & VBR & DualRead & Token & Paired $\Delta$\\
\midrule
Qwen3VL-2B & \ensuremath{.584_{\pm\!.020}} & \ensuremath{.532_{\pm\!.014}} & \ensuremath{.053_{\pm\!.007}} & \ensuremath{109.7_{\pm\!3.1}} & \ensuremath{59.7_{\pm\!1.5}}/\ensuremath{47.7_{\pm\!2.1}} & \ensuremath{.248_{\pm\!.007}} & \ensuremath{\mathbf{.784}_{\pm\!.009}} & \ensuremath{.516_{\pm\!.031}} & $+.269_{\pm.022}$\\
MedGemma1.5-4B & \ensuremath{.600_{\pm\!.010}} & \ensuremath{.566_{\pm\!.018}} & \ensuremath{.034_{\pm\!.015}} & \ensuremath{111.0_{\pm\!8.7}} & \ensuremath{59.7_{\pm\!2.5}}/\ensuremath{48.0_{\pm\!3.6}} & \ensuremath{.251_{\pm\!.020}} & \ensuremath{\mathbf{.819}_{\pm\!.030}} & \ensuremath{.503_{\pm\!.005}} & $+.316_{\pm.032}$\\
\bottomrule
\end{tabular*}

\vspace{4pt}
\textit{(c) Natural/strict decisive-set audit}\par\smallskip
\begin{tabular*}{\textwidth}{@{\extracolsep{\fill}}lrrrrrr@{}}
\toprule
& \multicolumn{3}{c}{Qwen3VL-2B} & \multicolumn{3}{c}{MedGemma1.5-4B}\\
\cmidrule(lr){2-4}\cmidrule(lr){5-7}
Actor & $|\mathcal E|$ & Strict & Retained & $|\mathcal E|$ & Strict & Retained\\
\midrule
RLVR & \ensuremath{109.7_{\pm\!3.1}} & \ensuremath{75.0_{\pm\!5.3}} & $65$--$72\%$ & \ensuremath{111.0_{\pm\!8.7}} & \ensuremath{77.7_{\pm\!8.5}} & $67$--$72\%$\\
RLCR & \ensuremath{101.0_{\pm\!3.0}} & \ensuremath{79.0_{\pm\!2.0}} & $74$--$83\%$ & \ensuremath{102.7_{\pm\!8.1}} & \ensuremath{67.0_{\pm\!1.7}} & $61$--$69\%$\\
DCPO & \ensuremath{110.3_{\pm\!8.5}} & \ensuremath{81.0_{\pm\!7.0}} & $73$--$74\%$ & \ensuremath{96.7_{\pm\!8.1}} & \ensuremath{61.7_{\pm\!4.5}} & $62$--$67\%$\\
VL-Cal. & \ensuremath{94.0_{\pm\!5.3}} & \ensuremath{66.3_{\pm\!2.1}} & $69$--$73\%$ & \ensuremath{75.0_{\pm\!3.0}} & \ensuremath{54.0_{\pm\!1.0}} & $71$--$74\%$\\
\bottomrule
\end{tabular*}
\caption{Grounding evidence: (a) OOD CCG-AUC, (b) paired SLAKE audit, and (c) eligibility. Bold marks the best confidence score.}
\label{tab:ccg_ood}\label{tab:ccg}\label{tab:ccg_denom}
\end{table*}

\section{Reproducibility and Scope}
\label{app:reproducibility}

For each seed, the order is greedy generation, deterministic grading, one
teacher-forced replay, reader fitting on SLAKE-train fitting images,
temperature fitting on disjoint SLAKE-train calibration images, and frozen
evaluation on ID/OOD manifests. Two frozen aggregate records back the reported
numbers: one holding the method $\times$ domain $\times$ seed matrix behind the
OOD6 averages, and one holding the three-domain CCG decomposition. Both are
released with the code. Readers are actor-specific and
require internal-state access. OOD temperatures remain frozen. CCG is
conditional on an actor-defined decisive set, is underpowered below 30
independent source/donor images, and does not measure pixel localization or
clinical causality.

\end{document}